\documentclass[acmtog]{acmart}

\usepackage{graphicx}
\usepackage{amsmath}
\usepackage{booktabs}
\usepackage{multirow}
\usepackage{bm}

\usepackage{overpic}
\usepackage{enumitem} 
\usepackage{overpic} 
\usepackage{color}
\usepackage{xcolor}

\definecolor{turquoise}{cmyk}{0.65,0,0.1,0.3}
\definecolor{purple}{rgb}{0.65,0,0.65}
\definecolor{dark_green}{rgb}{0, 0.5, 0}
\definecolor{orange}{rgb}{0.8, 0.6, 0.2}
\definecolor{red}{rgb}{0.8, 0.2, 0.2}
\definecolor{darkred}{rgb}{0.6, 0.1, 0.05}
\definecolor{blueish}{rgb}{0.0, 0.3, .6}
\definecolor{light_gray}{rgb}{0.7, 0.7, .7}
\definecolor{pink}{rgb}{1, 0, 1}
\definecolor{greyblue}{rgb}{0.25, 0.25, 1}

\usepackage{blindtext}

\makeatletter
\DeclareRobustCommand\onedot{\futurelet\@let@token\@onedot}
\def\@onedot{\ifx\@let@token.\else.\null\fi\xspace}

\def\eg{\emph{e.g}\onedot} 
\def\ie{\emph{i.e}\onedot}

\makeatother

\AtBeginDocument{%
  \providecommand\BibTeX{{%
    \normalfont B\kern-0.5em{\scshape i\kern-0.25em b}\kern-0.8em\TeX}}}

\setcopyright{none}

\setcopyright{acmcopyright}\acmJournal{TOG}
\acmYear{2022}\acmVolume{41}\acmNumber{4}\acmArticle{56}\acmMonth{7} \acmDOI{10.1145/3528223.3530177}

\acmSubmissionID{730}

\begin{document}

\title{NeROIC: Neural Rendering of Objects from Online Image Collections}


\author{Zhengfei Kuang}
\authornote{This work was performed while the author was an intern at Snap Inc.}
\email{zkuang@usc.edu}
\affiliation{%
  \institution{University of Southern California}
  \country{USA}
}

\author{Kyle Olszewski}
\email{olszewski.kyle@gmail.com}
\orcid{0000-0001-8775-6879}
\affiliation{%
  \institution{Snap Inc.}
  \country{USA}
}

\author{Menglei Chai}
\email{cmlatsim@gmail.com}
\affiliation{%
  \institution{Snap Inc.}
  \country{USA}
}

\author{Zeng Huang}
\email{frequencyhzs@gmail.com}
\affiliation{%
  \institution{Snap Inc.}
  \country{USA}
}

\author{Panos Achlioptas}
\email{panos@cs.stanford.edu}
\affiliation{%
  \institution{Snap Inc.}
  \country{USA}
}

\author{Sergey Tulyakov}
\email{sergey.tulyakov@unitn.it}
\affiliation{%
  \institution{Snap Inc.}
  \country{USA}
}
\renewcommand{\shortauthors}{Zhengfei Kuang, Kyle Olszewski, Menglei Chai, Zeng Huang, Panos Achlioptas, Sergey Tulyakov}

\begin{abstract}

We present a novel method to acquire object representations from online image collections, capturing high-quality geometry and material properties of arbitrary objects from photographs with varying cameras, illumination, and backgrounds.
This enables various object-centric rendering applications such as novel-view synthesis, relighting, and harmonized background composition from challenging in-the-wild input.
Using a multi-stage approach extending neural radiance fields, we first infer the surface geometry and refine the coarsely estimated initial camera parameters, while leveraging coarse foreground object masks to improve the training efficiency and geometry quality.
We also introduce a robust normal estimation technique which eliminates the effect of geometric noise while retaining crucial details. Lastly, we extract surface material properties and ambient illumination, represented in spherical harmonics with extensions that handle transient elements, \eg sharp shadows. The union of these components results in a highly modular and efficient object acquisition framework. Extensive evaluations and comparisons demonstrate the advantages of our approach in capturing high-quality geometry and appearance properties useful for rendering applications.

\end{abstract}


\begin{teaserfigure}
    
\begin{center}

    
\begin{tabular}{c}
    \begin{tabular}{ccc}

    \end{tabular} \\
    
    \setlength{\tabcolsep}{1.0pt}
    \begin{tabular}{cccccc}        
        
        \includegraphics[width=0.18\textwidth]{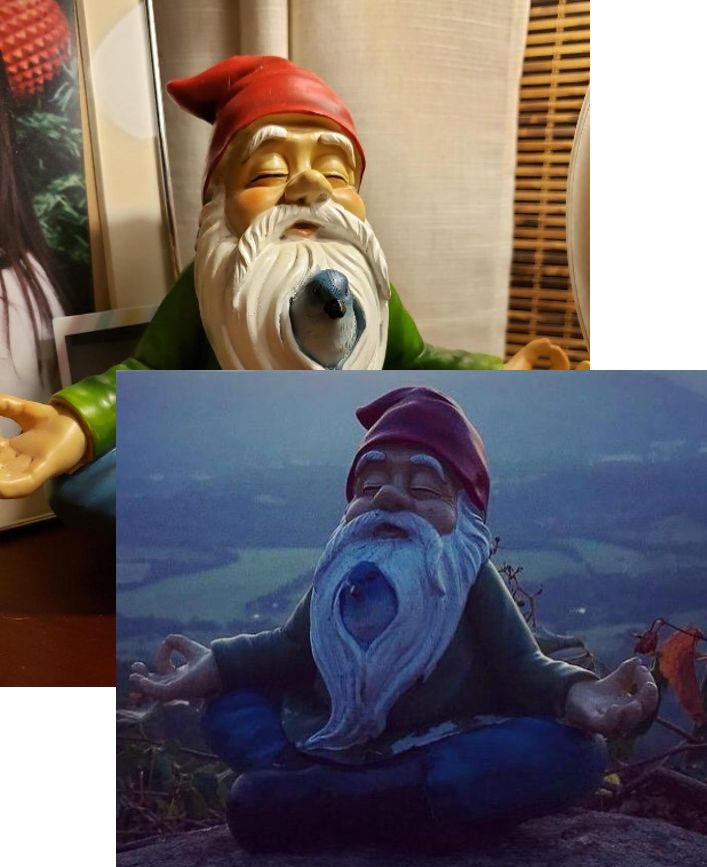} &
        \includegraphics[width=0.20\textwidth]{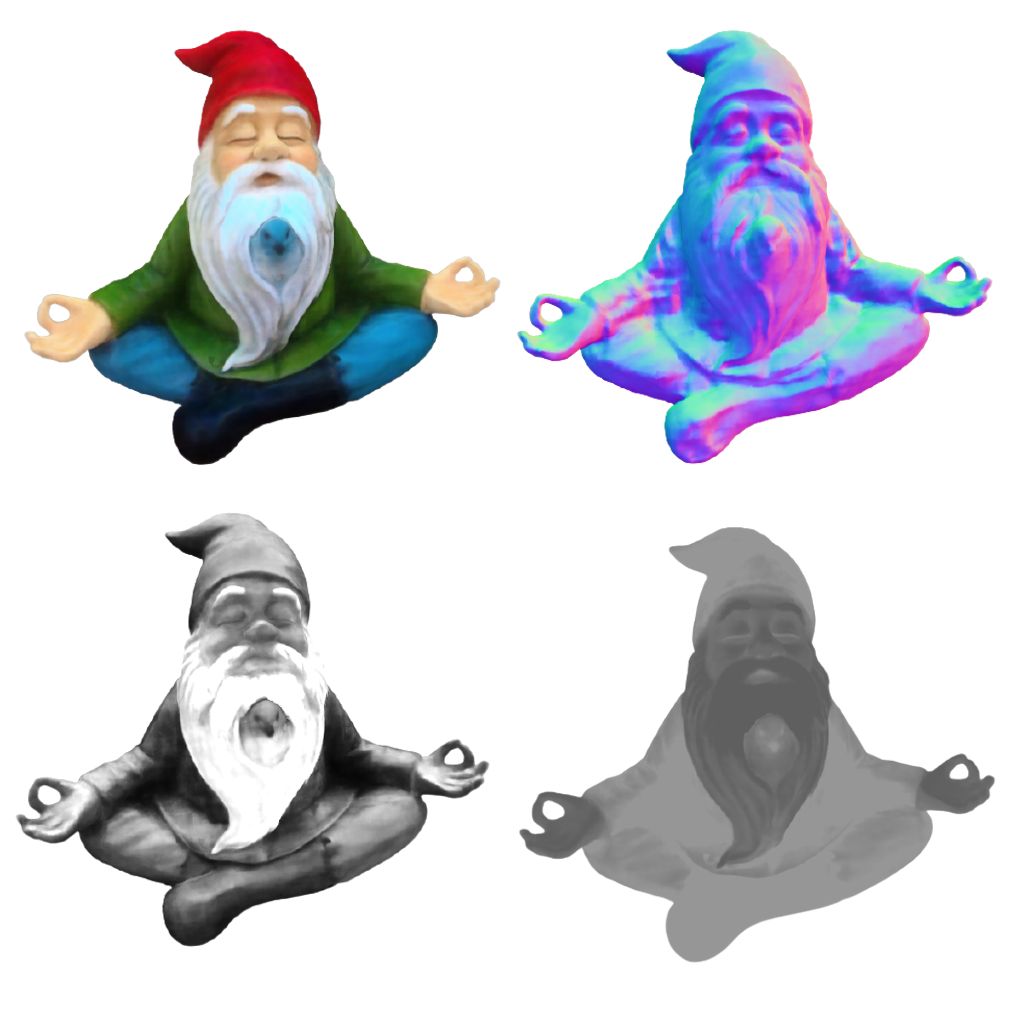} &
        \includegraphics[height=0.20\textwidth]{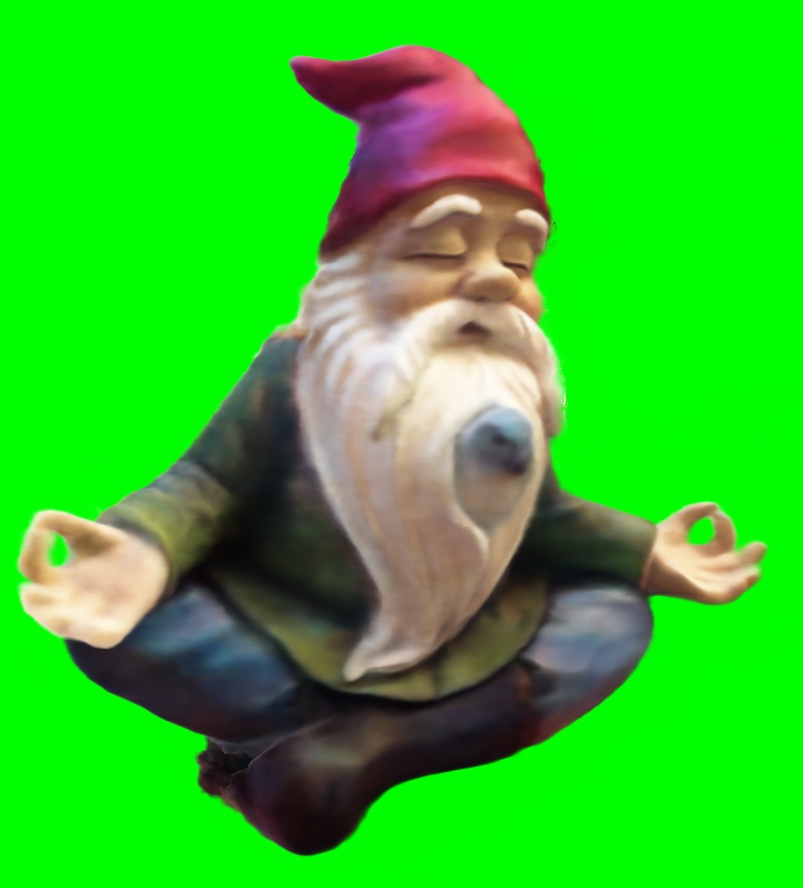} &
        \includegraphics[height=0.20\textwidth]{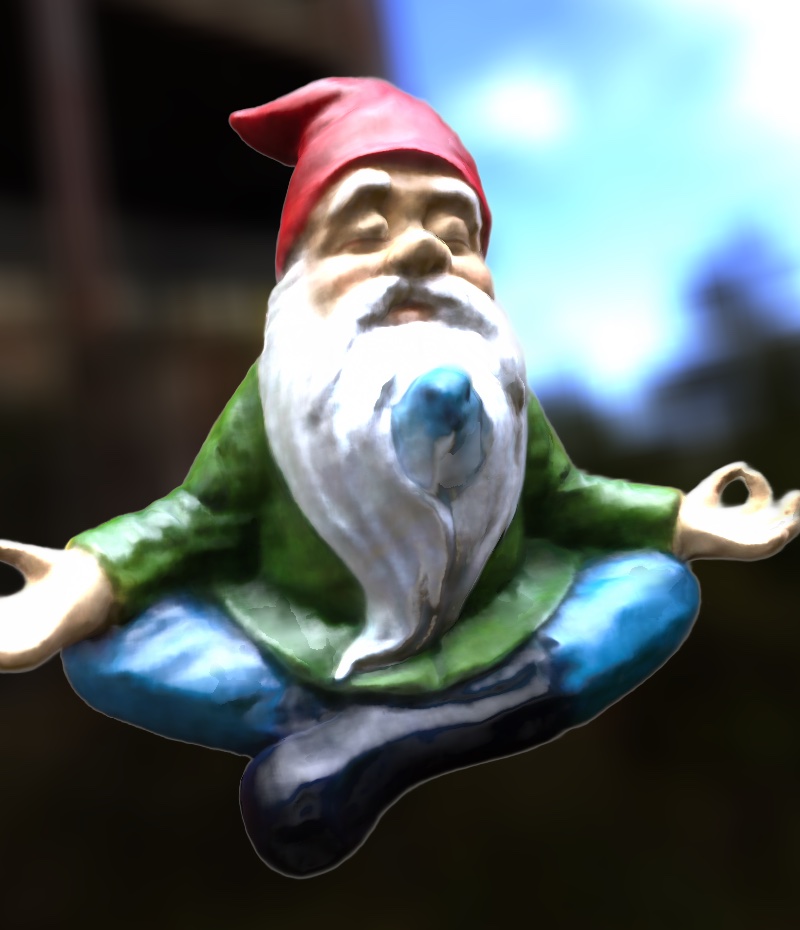} &
        \includegraphics[height=0.20\textwidth]{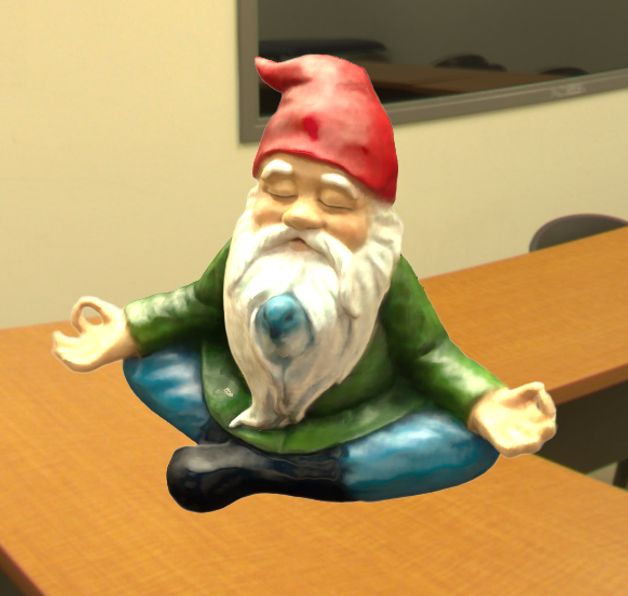} \\
        {Online Images} & {Outputs} &
        Novel View Synthesis & Relighting & Composition \\
    \end{tabular}
\end{tabular}

\end{center}

\caption{
\textbf{Our Object Capture Results from Online Images.}
Our modular NeRF-based approach requires only sparse, coarsely segmented images depicting an object captured under widely varying conditions (1st column).
We first infer the geometry as a density field using neural rendering, and then compute the object's surface material properties and per-image lighting conditions (2nd column).
Our model not only can synthesize novel views, but can also relight and composite the captured object in novel environments and lighting conditions (3-5th column).
}
\label{fig:teaser}
\end{teaserfigure}

\maketitle

\section{Introduction}
\label{sec:intro}

Numerous collections of images featuring identical objects, \eg furniture, toys, vehicles, can be found online on shopping websites or through a simple image search.
The ability to isolate these objects from their surroundings and capture high-fidelity structure and appearance is highly desired, as it would enable applications such as digitizing an object from the images and blending it into a new background.
However, individual images of the objects in these collections are typically captured in highly variable backgrounds, illumination conditions, and camera parameters, making object digitization approaches specifically designed for data from controlled 
environments unsuitable for such an \textit{in-the-wild} setup. In this work, we seek to address this challenge by developing an approach for capturing and re-rendering objects from unconstrained image collections by extending the latest advances in neural object rendering.

Among the more notable recent works using neural field is the Neural Radiance Fields (NeRF) model~\cite{mildenhall2020nerf}, which learns to represent the local opacity and view-dependent radiance of a static scene from sparse calibrated images, allowing high-quality novel view synthesis (NVS). While substantial progress has been made to improve the quality and capabilities of NeRF (\eg moving or non-rigid content~\cite{DBLP:conf/cvpr/PumarolaCPM21,DBLP:journals/tog/ParkSHBBGMS21,DBLP:conf/cvpr/LiNSW21,DBLP:conf/cvpr/XianHK021}), some non-trivial requirements still remain -- to synthesize novel views of an object the background and illumination conditions should be seen and fixed, and the multi-view images or video sequences should be captured in a single session.

Recently, several works~\cite{guo2020osf,DBLP:conf/cvpr/Martin-BruallaR21,DBLP:conf/iccv/BossBJBLL21,neuralpil,DBLP:journals/tog/ZhangSDDFB21,DBLP:conf/iccv/Yang0XLZB0C21} have extended NeRF and achieved impressive progress in decomposing the renderings of a scene into semantically meaningful components, including geometry, reflectance, material, and lighting, enabling a flexible interaction with any of these components, \eg relighting and swapping the background. Unfortunately, none of them built a comprehensive solution 
to work with the limitations of objects captured from real-world, in-the-wild image collections.
In this work, we propose \textbf{NeROIC}, a novel approach to \textbf{Ne}ural \textbf{R}endering of objects from \textbf{O}nline \textbf{I}mage \textbf{C}ollections.
Our object capture and rendering approach builds upon neural radiance fields with several key features that enable high-fidelity capture from sparse images captured under wildly different conditions, which is commonly seen in online image collections with individual images taken with varying lightings, cameras, environments, and poses.
The only expected annotation for each image is a rough foreground segmentation and coarsely estimated camera parameters, which crucially we can obtain in an unsupervised way from structure-from-motion frameworks such as COLMAP~\cite{colmap}.

Key to our learning-based method is the introduction of a modular approach, in which we first optimize a NeRF model to estimate the geometry and refine the camera parameters, and then infer the surface material properties and per-image lighting conditions that best explain the captured images.
The decoupling of these stages allows us to use the depth information from the first stage to do more efficient ray sampling in the second stage, which improves material and lighting estimation quality and training efficiency. Furthermore, due to the modularity of our approach we can also separately exploit the surface normals initialized from the geometry in the first stage, and innovate with a new normal extraction layer that 
enhances the accuracy of acquiring materials of the underlying object.
An overview of our approach is shown in  Fig.~\ref{fig:overview} (b).

To evaluate our approach, we create several in-the-wild object datasets, including images captured by ourselves in varying environments, as well as images of objects collected from online resources.
The comparisons with state-of-the-art alternatives, in these challenging setups, indicate that our approach outperforms the alternatives qualitatively and quantitatively, while still maintaining comparable training and inference efficiency. Fig.~\ref{fig:teaser} presents a set of example object capturing and application results by our approach.

In summary, our main contributions are:
\begin{itemize}[leftmargin=*]
    \item A novel, modular pipeline for inferring geometric and material properties from objects captured under varying conditions, using only sparse images, foreground masks, and coarse camera poses as additional input,
    \item A new multi-stage architecture where we first extract the geometry and refine the input camera parameters, and then infer the object's material properties, which we show is robust to unrestricted inputs, 
    \item A new method for estimating normals from neural radiance fields that enables us to better estimate material properties and relight objects than more standard alternative techniques,
    \item Datasets containing images of objects captured in varying and challenging environments and conditions,
    \item Extensive evaluations, comparisons and results using these and other established datasets demonstrating the state-of-the-art results obtained by our approach.
\end{itemize}

Our code, pre-trained models, and training datasets are released at \url{https://formyfamily.github.io/NeROIC/}.


\section{Related Work}
\label{sec:related}
\paragraph{Neural Rendering for Novel View Synthesis}
One of the more recent advances in novel view synthesis is NeRF~\cite{mildenhall2020nerf}. A set of multilayer perceptrons (MLPs) are used to infer the opacity and radiance for each point and outgoing direction in the scene by sampling camera rays and learning to generate the corresponding pixel color using volume rendering techniques, allowing for high-quality interpolation between training images.
However, this framework requires well-calibrated multi-view datasets of static scenes as input, with no variation in the scene content and lighting conditions.
Many subsequent works build upon this framework to address these and other issues.
NeRF-{}-~\cite{Wang21arxiv_nerfmm}, SCNeRF~\cite{DBLP:conf/iccv/JeongACACP21} and BARF~\cite{DBLP:conf/iccv/LinM0L21} infer the camera parameters while learning a neural radiance field, to allow for novel view synthesis when these parameters are unknown.
iNeRF~\cite{DBLP:conf/iros/LinFBRIL21} estimate poses by inverting a trained neural radiance field to render the input images.
Other works focus on improving the training or inference performance and computational efficiency~\cite{DBLP:conf/iccv/YuLT0NK21,Liu20neurips_sparse_nerf,DBLP:conf/iccv/ReiserPL021,DBLP:conf/cvpr/WangWGSZBMSF21,DBLP:conf/cvpr/LindellMW21,DBLP:journals/cgf/NeffSPKMCKS21}).
Related approaches~\cite{Kellnhofer:2021:nlr,bergman2021metanlr} use a signed-distance function to represent a surface that can be extracted as a mesh for fast rendering and novel view synthesis.
However, these works only display high-quality results for a limited range of interpolated views, and do not perform the level of material decomposition and surface reconstruction needed for high-quality relighting and reconstruction.


\paragraph{Learning from Online Image Collections}
Online image collections have been used for various applications, such as reconstructing the shape and appearance of buildings~\cite{snavely2008modeling} or human faces~\cite{6126439,liang2016head}.
However, such approaches typically require many available photographs, making them applicable only to landmarks or celebrities, and are designed explicitly to work with specific subjects with domain features, rather than arbitrary objects.
Recently, neural rendering has been combined with the use of generative adversarial networks~\cite{goodfellow2014generative} to allow for generatively sampling different objects within a category and rendering novel views~\cite{DBLP:journals/corr/abs-1904-01326,DBLP:conf/cvpr/Niemeyer021,Schwarz20neurips_graf}, using only a single image of each training object.
However, such approaches do not allow for rendering novel views of a target object, \eg from one or more images, with controllable shape, appearance, and environmental conditions, and the quality of the sampled images varies.
Other approaches learn pose, shape, and texture from images for certain categories of objects~\cite{meshry2019neural,jang2021codenerf}, or interpolation, view synthesis, and segmentation of sampled category instances~\cite{xie2021fignerf}.
However, none of these approaches allow for the level of structure and material decomposition suitable for high-fidelity rendering and relighting.

\paragraph{Image Content Decomposition and Relighting}
Many recent works focus on decomposing the lighting condition and intrinsic properties of the objects from the training images. 
NeRF-in-the-Wild~\cite{DBLP:conf/cvpr/Martin-BruallaR21} learns to render large-scale scenes from images captured at different times by omitting inconsistent and temporary content, such as passersby, and implicitly representing lighting conditions as appearance features that can be interpolated.
However, this approach does not fully decompose the scene into geometric and material properties for arbitrary lighting variations, and is not designed to address challenging cases such as extracting isolated objects from their surroundings.
On the other hand, many works including Neural Reflectance Field~\cite{Bi20arxiv_neural_reflection_fields}, NeRFactor~\cite{DBLP:journals/tog/ZhangSDDFB21}, NeRV~\cite{DBLP:conf/cvpr/SrinivasanDZTMB21}, and PhySG~\cite{physg} combine NeRF with physically-based rendering techniques, and estimate various material properties of the target object. However, all of these works require well-conditioned or known lighting, and are not adaptive to input images from unknown arbitrary environments. Some recent works, i.e. NeRD~\cite{DBLP:conf/iccv/BossBJBLL21},  Neural-PIL~\cite{neuralpil},  NeRS~\cite{ners} relax the constraint of dataset, but they still requires restrictions on the inputs, such as known exposure and white balancing parameters~\cite{DBLP:conf/iccv/BossBJBLL21}, data from the same source~\cite{ners}. Most importantly, all of these approaches are inevitably vulnerable to inputs with complex shading, e.g. sharp shadows and mirror-like reflections, since they only consist of one physical-based renderer which is relatively simple. While we do not take the claim to learn how to fit those shadings in our method, in our work we introduce a transient component based on~\citet{DBLP:conf/cvpr/Martin-BruallaR21} to identify and disentangle it from environment lighting, thus acquiring unbiased material properties of the object.
Another recent work on neural decomposition, NeRF-OSR~\cite{DBLP:journals/corr/abs-2112-05140} uses Spherical Harmonic coefficients to represent environmental lighting as in our approach, but is designed for outdoor scene reconstruction (\eg, buildings).  To the best of our knowledge, we are the first NeRF-based method to infer both geometry and material parameters of the target with \emph{fully unconstrained images from the internet.}

\section{Method}
\label{sec:method}

In this section, we outline our approach to object-centric aggregation.
We first provide an overview of the approach (Sec.~\ref{sec:method_overview}), followed by a description of the neural radiance fields framework we extend in our method (Sec.~\ref{sec:method_nerf}).

\begin{figure*}[t]
\begin{center}
\setlength{\tabcolsep}{1.0pt}
\begin{tabular}{c|c|c}
\includegraphics[height=0.48\columnwidth]{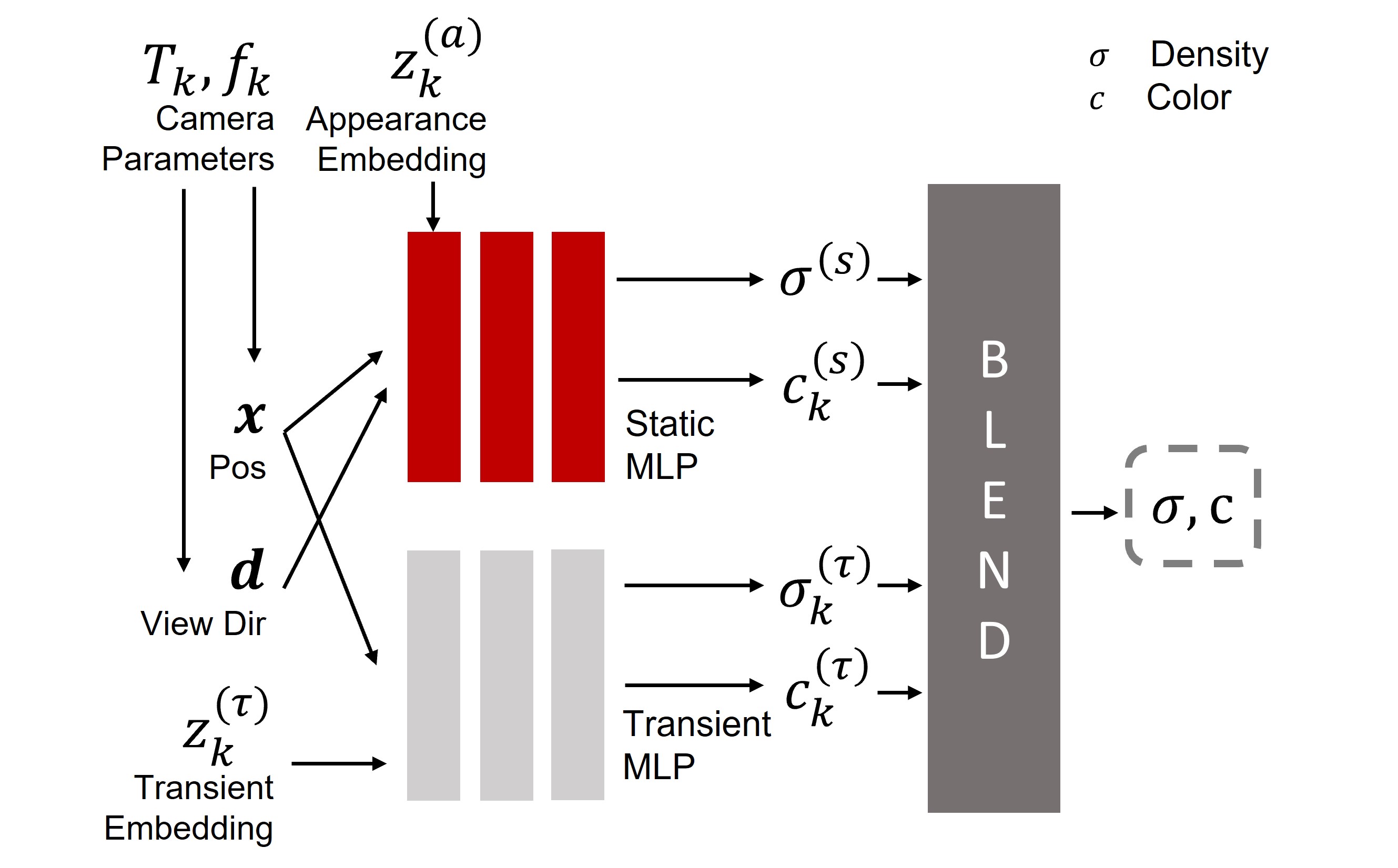} \hspace{-6pt} &
\includegraphics[height=0.48\columnwidth]{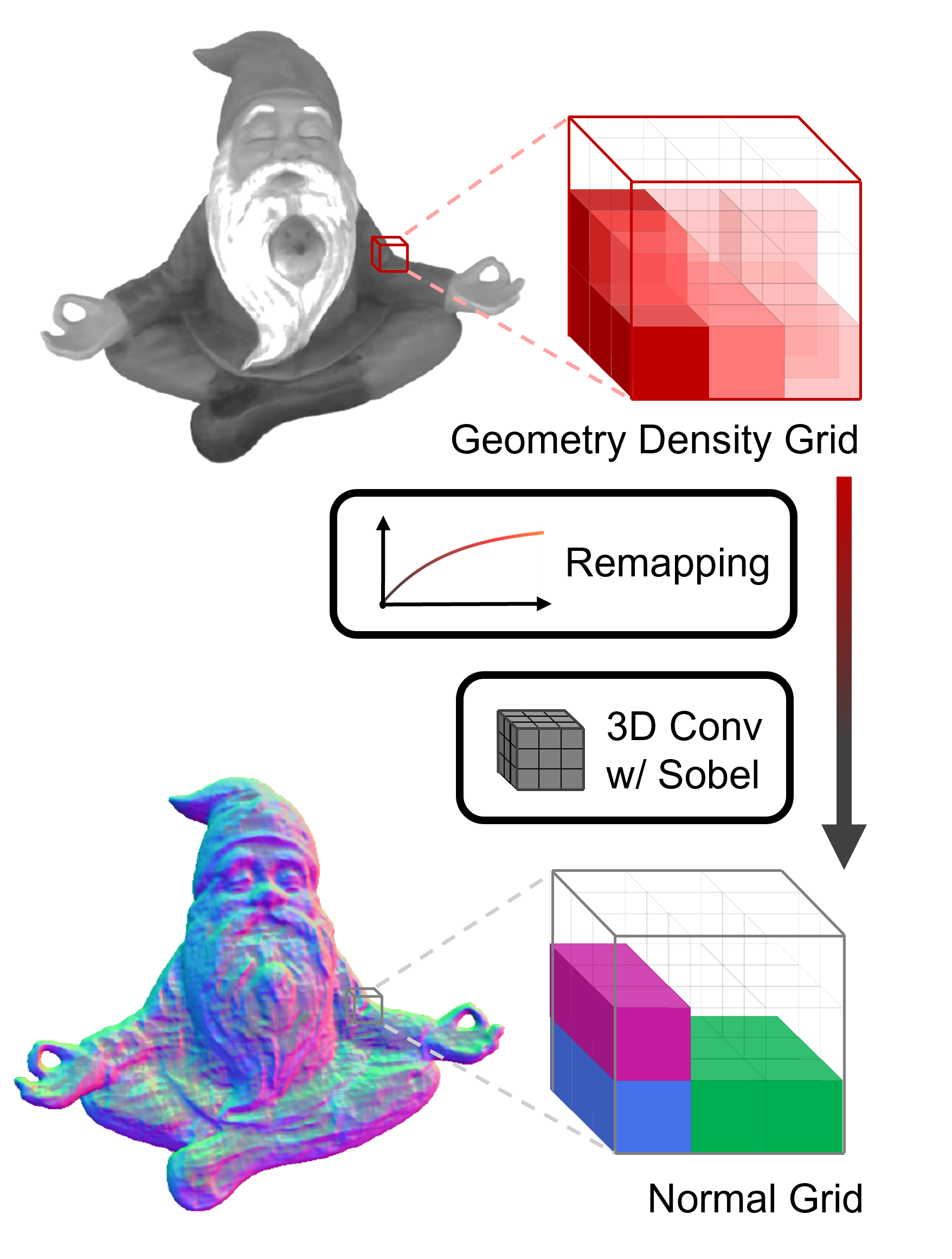} &
\hspace{-6pt} \includegraphics[height=0.48\columnwidth]{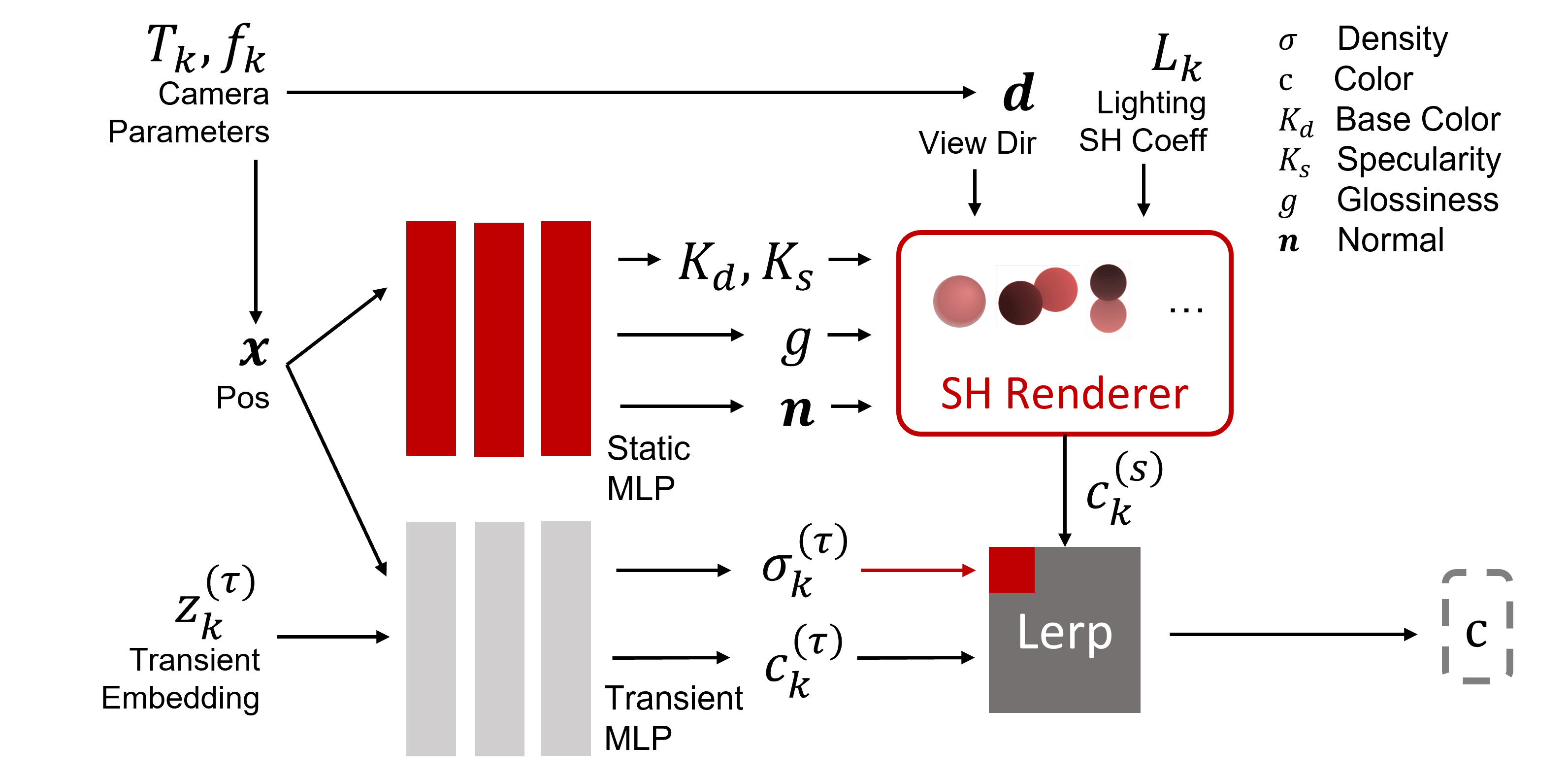} \\

\footnotesize(a) Geometry Network & \hspace{6pt} \footnotesize(b)   Normal Extraction \hspace{6pt} & \footnotesize(c) Rendering Network
\end{tabular}
\end{center}
\caption{
\textbf{Overview of Our Approach.}
Given a set of coarsely calibrated images and corresponding foreground masks, our geometry network computes a neural radiance field with both static and transient components, and refines the camera parameters (a).
Our grid-based normal extraction layer then estimates the surface normals from the learned density field (b).
Finally, we fix the geometry of the object and use the estimated normals as supervision in our rendering network, in which we infer the lighting conditions (represented as spherical harmonics coefficients), surface material properties (using the Phong rendering model), and high-quality surface normals (c). 
}
\label{fig:overview}
\end{figure*}

\subsection{Overview}
\label{sec:method_overview}

Fig.~\ref{fig:overview} provides an overview of our approach.
The inputs are a sparse collection of images $\mathcal{I}_k: [0,1]^2 \rightarrow [0,1]^3$ depicting an object (or instances of an identical object) under varying conditions, and a set of foreground masks $\mathcal{F}_k: [0,1]^2 \rightarrow \{0,1\}$ defining the region of the object, where $1\leq k\leq N$. During the first stage, we estimate the geometry of the object by learning a density field indicating where there is physical content (Sec.~\ref{sec:method_first_stage}).
During this stage, we also learn both static and transient radiance values to allow for image-based supervision, but do not fully decompose this information into material and lighting properties.
We also optimize the pose and intrinsic parameters of the cameras to refine the coarse estimates provided as input. 

In the second stage, we fix the learned geometry and optimize the surface material and lighting parameters needed to re-render the object in arbitrary illumination conditions (Sec.~\ref{sec:method_second_stage}).
During this stage, we use the estimated distance from the camera to the object surface to improve our point sampling along the camera rays. 
We also optimize the surface normals, which improves on the coarse estimates that are obtained from our density field (Sec.~\ref{sec:method_normal_opt}).

\subsection{Preliminaries}
\label{sec:method_nerf}

In Neural Radiance Fields (NeRF)~\cite{mildenhall2020nerf}, a set of networks are trained to infer radiance and density for arbitrary 3D points, and generate images from novel viewpoints using volumetric rendering. Specifically, it employs two MLP functions: a density function $\sigma(\bm{x}):\mathbb{R}^3\rightarrow\mathbb{R}^+$ and a color function $c(\bm{x}):\mathbb{R}^3\rightarrow[0,1]^3$. For each ray $\bm{r}=(\bm{r}_o,\bm{r}_d)$ emitted from the camera origin $\bm{r}_o$ in direction $\bm{r}_d$, NeRF samples $N_p$ 3D points along the ray $\bm{x}_i=\bm{r}_o+d_i\bm{r}_d$ ($0\leq i\leq N_p$), and integrates the pixel color as follows:
\begin{equation}
    C(\bm{r})=\sum_{i=1}^{N_p}\alpha_i(1-w_i)c(\bm{x}_i),
    \label{eq:NeRF}
\end{equation}
where 
$w_i=\exp(-(d_i-d_{i-1})\sigma(\bm{x}_i))$ represents the transmittance of the ray segment between sample points $\bm{x}_{i-1}$ and $\bm{x}_{i}$, and $\alpha_i=\prod_{j=1}^{i-1}w_i$ is the ray attenuation from the origin $\bm{r}_o$ to the sample point $\bm{x}_i$. In addition to the volumetric rendering function, NeRF also introduces an adaptive coarse-to-fine pipeline which uses a coarse model to guide the 3D point sampling of the fine model.

\subsection{Geometry Networks}
\label{sec:method_first_stage}


In our first stage, we seek to reconstruct the geometry of the target object depicted in our image collection.
This, however, is made more challenging due to the varying lighting environments, transient conditions \eg sharp shadows, varying camera parameters, and coarse camera poses and intrinsics caused by the lack of background context required for accurate camera calibration.
Inspired by~\citet{DBLP:conf/cvpr/Martin-BruallaR21}, we employ a pipeline designed to make use of images captured under different conditions, and introduce additional designs to account for the challenging task of aggregating an object representation solely from the isolated foreground region.

Thus, we employ a two-branch pipeline which handles transient and static content separately, and assigns unique embedding vectors $\bm{z}^{(\tau)}_k$ and $\bm{z}^{(a)}_k$ to each image to represent the transient geometry and changing lighting.
Our model for this stage thus consists of four functions instead of two: $\sigma^{(s)}(\bm{x}),\sigma^{(\tau)}_k(\bm{x})$, and $c^{(s)}_k(\bm{x}),c^{(\tau)}_k(\bm{x})$.
The volumetric rendering function in~\citet{mildenhall2020nerf} is re-formulated as:
\begin{equation}
\begin{gathered}
C_k(\bm{r}) = \\ \sum_{i=1}^{N_p} \alpha_{ki} ( (1-w^{(s)}_{ki}) c^{(s)}_k(\bm{x}_i) + (1-w^{(\tau)}_{ki}) c^{(\tau)}_k(\bm{x}_i)),
\end{gathered}
\label{eq:nerfw}
\end{equation}
where $w^{(s,\tau)}_{ki} = \exp(-(d_i-d_{i-1})\sigma^{(s,\tau)}_k(\bm{x}_i))$,
and $\alpha_{ki} = \prod_{j=1}^{i-1} w^{(s)}_{kj} w^{(\tau)}_{kj}$. We also adopt the Bayesian learning framework of~\citet{DBLP:conf/nips/KendallG17}, predicting an uncertainty $\beta_k(x)$ for transient geometry when accounting for the image reconstruction loss.

Eq.~\ref{eq:nerfw} serves as the rendering
function when training this network.
As in~\citet{DBLP:conf/cvpr/Martin-BruallaR21}, we use a color reconstruction loss $\mathcal{L}_{c}$ incorporated with $\beta_k$, and a transient regularity loss $\mathcal{L}_{tr}$.~\footnote{For details on these losses and their use, please see the supplementary document.}
However, to accurately capture the geometry corresponding to our target object, we found it essential to incorporate additional losses designed for our particular use case.

\paragraph{Silhouette Loss}
We use the input foreground masks to help the networks focus on the object inside the silhouette, thus preventing ambiguous geometry from images with varying backgrounds. While we mask out the background in each image and replace it with pure white, a naive approach will still fail to discriminate the object from the background, thus producing white artifacts around the object and occluding it in novel views. To avoid this issue, we introduce a silhouette loss $\mathcal{L}_{sil}$, defined by the binary cross entropy (BCE) between the predicted ray attenuation $\alpha_k$ and the ground-truth foreground mask $\mathcal{F}_k$ to guide the geometry learning process.
As seen in the ablation study in Tab.~\ref{tab:ablation_1}, the silhouette loss significantly improves our results on testing data.

\paragraph{Adaptive Sampling}
We also introduce an adaptive sampling strategy in our model using these masks.
At the beginning of every training epoch, we randomly drop out part of the background rays from the training set, to ensure that the ratio of the foreground rays is above $1/3$.
This seemingly simple strategy significantly increases the training efficiency, and balances the silhouette loss and prevents $\alpha_{k}$ from converging to a constant.
Our ablation study in Tab.~\ref{tab:ablation_1} demonstrates that without this adaptive sampling, the model produces much worse results during testing.

\begin{figure}[t]
\begin{center}
\newcommand{\img}[1]{\includegraphics[height=0.12\textwidth]{#1}}
\newcommand{\imgl}[1]{\includegraphics[width=0.15\textwidth]{#1}}
\begin{tabular}{ccc}
\imgl{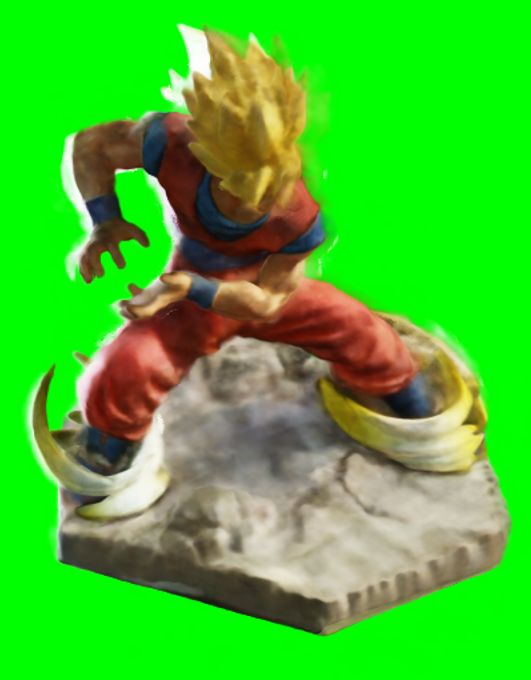} & 
 \imgl{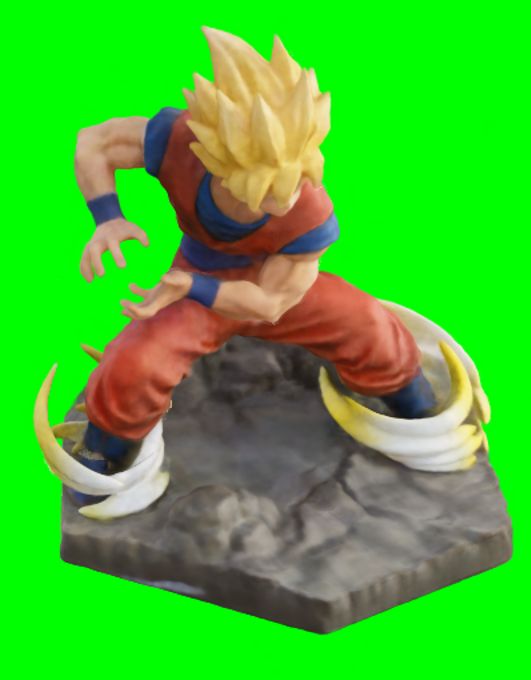} & 
\imgl{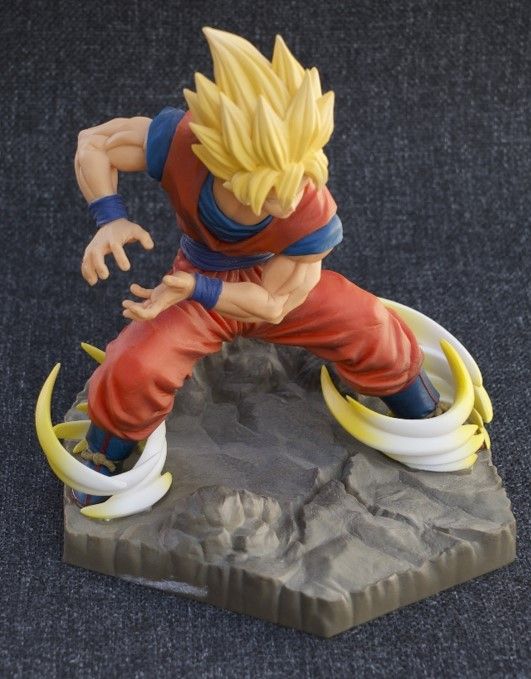} \\

\footnotesize w/o Cam Opt & \footnotesize w/ Cam Opt & \footnotesize GT  \\

\end{tabular}
\end{center}
\caption{
\textbf{Comparison on Camera Optimization.} The model trained without camera optimization produces object geometry and color of poorer quality than the full model.}
\label{fig:ablation_camera}

\end{figure}

\paragraph{Camera Optimization}
While our input images come from multiple sources, the lack of a consistent background leads to poor camera pose registration.
In practice, though we use COLMAP~\cite{colmap} on images with the backgrounds removed, the poses for some objects are still inaccurate, as seen in Fig.~\ref{fig:ablation_camera}.
To address this issue, we jointly optimize the camera poses during training, in a manner similar to~\citet{Wang21arxiv_nerfmm}.
More specifically, we incorporate camera parameters $(\delta_R,\delta_t,\delta_f)$ for rotation, translation, and focal length, respectively.
We use an axis-angle representation for rotation, while the others are in linear space.
We also add a regularity loss $\mathcal{L}_{cam}$ for the camera parameters, which is simply an L2 loss on these parameters.

\setlength{\tabcolsep}{1.5pt}

\begin{figure}[ht]
\begin{center}
\newcommand{\img}[1]{\includegraphics[width=0.18\textwidth]{#1}}
\newcommand{\smallimg}[1]{\includegraphics[width=0.11\textwidth]{#1}}
\newcommand{\middleimg}[1]{\includegraphics[width=0.15\textwidth]{#1}}
\begin{tabular}{c|c}

    \setlength{\tabcolsep}{2.0pt}
    \begin{tabular}{c}
        \middleimg{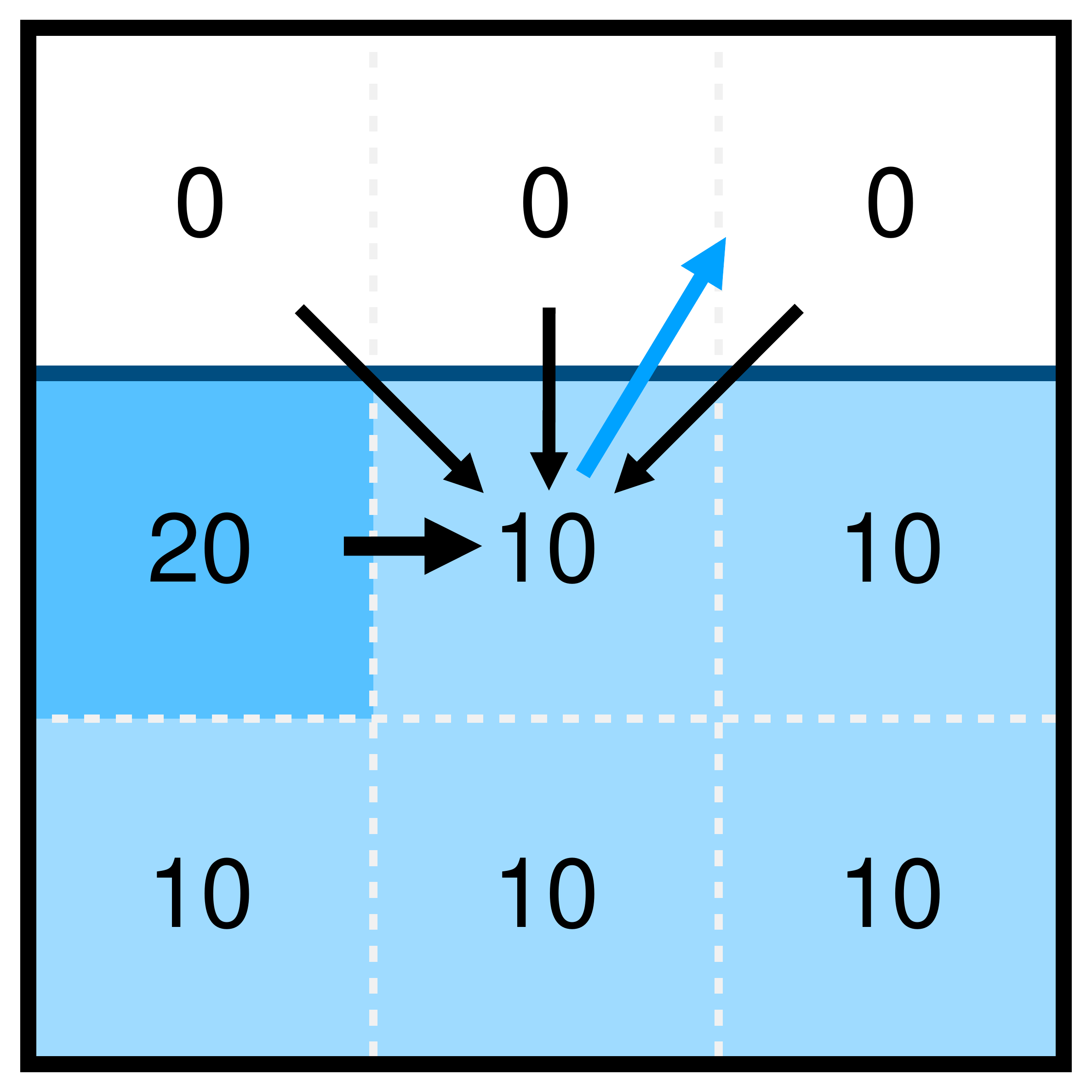}\\
        \footnotesize w/o Remap
        \\
        
        \middleimg{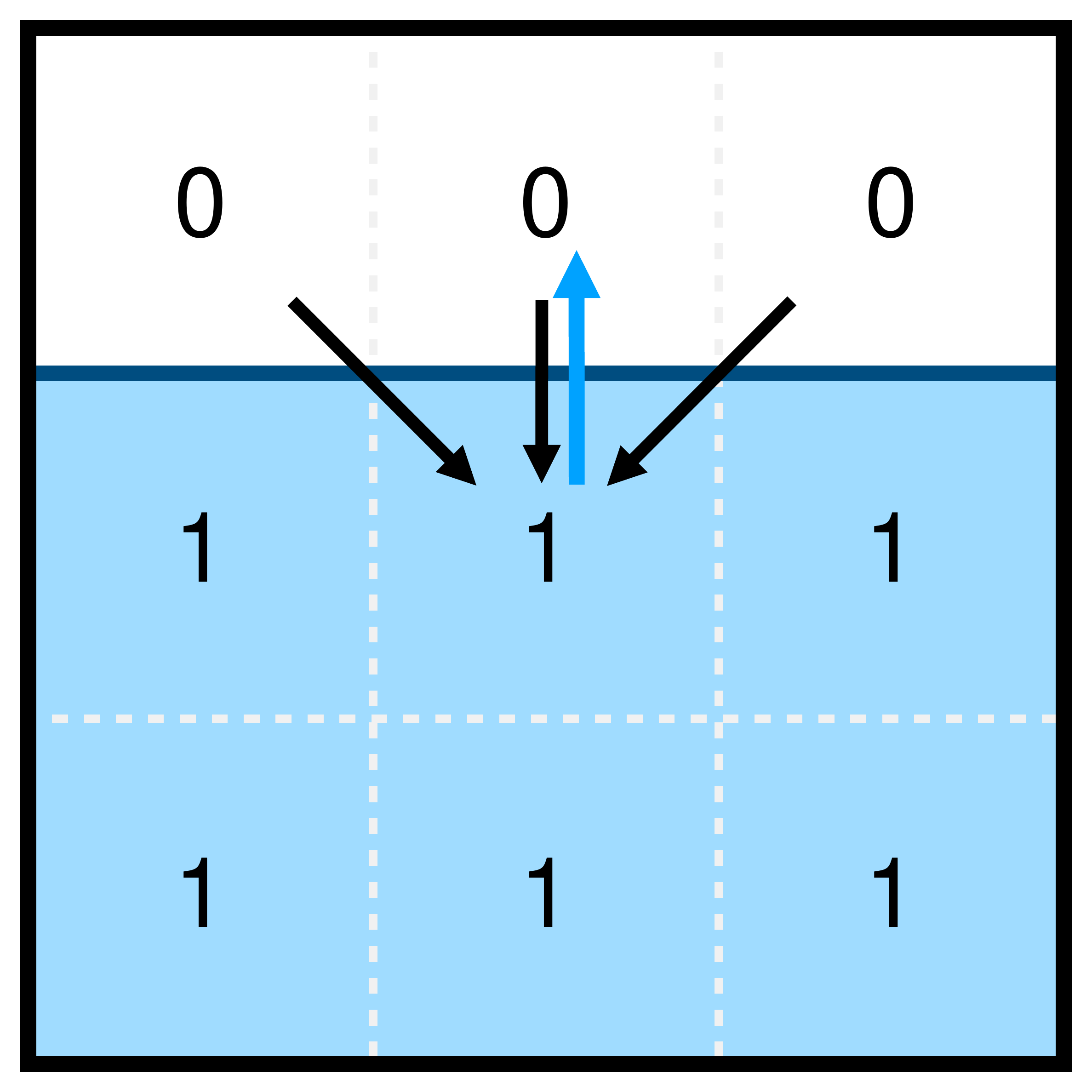}\\
        \footnotesize w/ Remap
        \\

    \end{tabular} & 

    \setlength{\tabcolsep}{1.0pt}
    \begin{tabular}{cccc}
        \smallimg{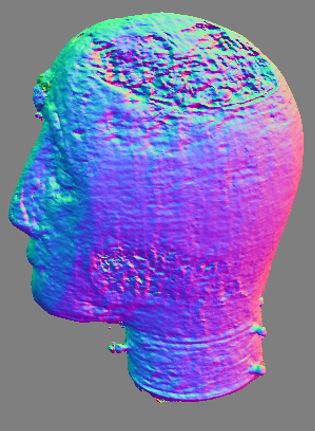} &
        \smallimg{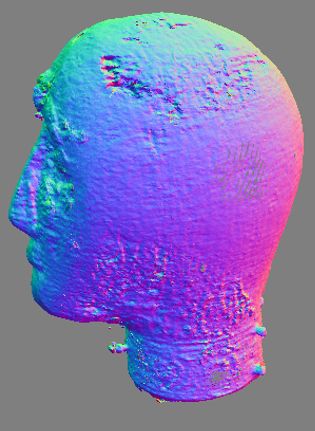} \\
        \footnotesize Original & \footnotesize Remapped \\
        \smallimg{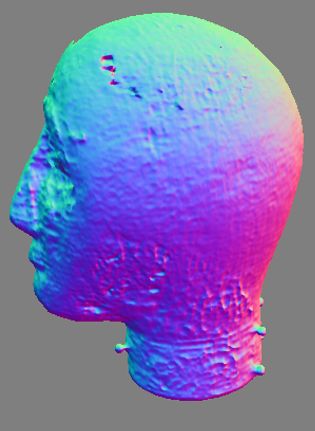} &
        \smallimg{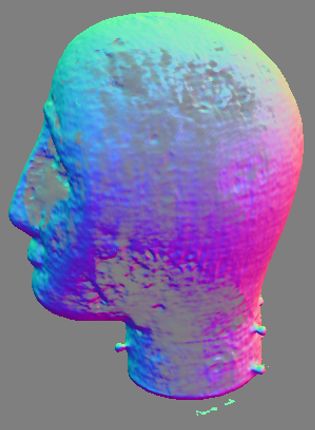} \\
        \footnotesize Ours & \footnotesize Ours w/ Conf
    \end{tabular} \\


\end{tabular}
\end{center}
\caption{\textbf{Analysis of Normal Extraction Layer.} On the left, while the gradient-based normal prediction (blue arrow) may be affected by noise in an unbounded density field, this effect can be alleviated by density remapping ($\lambda=1$ in this case). On the right, we show the estimated normals from the original density field (top left), remapped normals (top right), our normal extraction layer output (bottom left), and our result with confidence (bottom right).}
\label{fig:analysis_component}
\end{figure}

\setlength{\tabcolsep}{1.5pt}

As a summary, the final loss we use for this stage is:
\begin{equation}
\mathcal{L}_{geo} = \mathcal{L}_c + \lambda_{tr}\mathcal{L}_{tr} + \lambda_{sil}\mathcal{L}_{sil} + \lambda_{cam}\mathcal{L}_{cam},
\end{equation}
where the weights $\lambda_{tr}$, $\lambda_{sil}$, and $\lambda_{cam}$ are $0.01$, $0.1$, and $0.01$, respectively, in our experiments.





\subsection{Normal Extraction Layer}
\label{sec:method_normal_opt}
With the learned geometry from our first stage, we then extract the surface normals of the object as the supervision to the next stage, which helps reduce the ambiguity of the lighting and material estimation task. While many previous works~\cite{DBLP:journals/tog/ZhangSDDFB21,Bi20arxiv_neural_reflection_fields,DBLP:conf/iccv/BossBJBLL21} choose to use the gradient of the density function (i.e., $\nabla \sigma^{(s)}(\bm{x})$) as an approximation of normals, we find that this approach may produce incorrect results in certain areas, due to the challenging issues with unconstrained, real data (blurry images, varying lighting) that reduce the geometry quality and introduce noise into the density function. As explained in Fig.~\ref{fig:analysis_component}, this noise can drastically mislead the normal estimation without changing the surface shape itself. To resolve this, we propose a novel normal estimating pipeline based on the remapping of the density function and 3D convolution on a dense grid, which can produce smooth and accurate normals even with defective density.

We first calculate the bounding box of the object. To do so, we sparsely sample pixels of training images that are inside the foreground mask, and extract the expected surface-ray intersections for each ray, gathered as a point cloud. We directly compute the bounding box on it. After that, we discretize the bounding box into a $512^3$ dense grid and extract the density of each grid center. For a grid center $\bm{x}$, we remap its density value as:
\begin{equation}
    \sigma'_{\bm{x}} = \frac{1}{\lambda}(1-\exp(-\lambda\sigma_{\bm{x}})).
\end{equation}
This function remaps the density value from $[0,+\infty]$ to $[0,\frac{1}{\lambda}]$. The derivative gradually decays as the density value increases, which helps to filter out noise and obtain smoother predictions. $\lambda$ is a controllable parameter to adjust the sharpness of the normal. As $\lambda$ decreases, the derivative converges to a constant value of $1$, and thus the mapping function itself converges to the identity function. After remapping, we estimate the gradient of the density field $d\sigma'/d\mathbf{x}$ by applying a 3D convolution with a Sobel kernel $\mathcal{K}(\mathbf{x})=\mathbf{x} / ||\mathbf{x}||_2^2$ of size $5$ to the density grid.
 
Finally, we divide the convolution output $\bm{n}^{(g)}_{\bm{x}}=-\mathcal{K}(\sigma'_{\bm{x}})$ by $\max(1, ||\bm{n}^{(g)}_{\bm{x}}||_2^2)$, producing a normal supervision vector with length no larger than $1$. We treat its length as the confidence of the estimation, which becomes the weight of its supervising loss in the following stage. We show the results of each step in Fig.~\ref{fig:analysis_component}.

\subsection{Rendering Networks}
\label{sec:method_second_stage}

The purpose of our final stage is to estimate the lighting of each input image and the material properties of the object, given the geometry shape and surface normals from previous stages. Since extracting object materials in unknown lighting is highly ill-posed~\cite{sh-rendering,DBLP:journals/tog/ZhangSDDFB21}, we use low-order Spherical Harmonics (SH) to represent our lighting model and optimize its coefficients.
We use the standard Phong BRDF~\cite{phong1975illumination} to model the object material properties, which are controlled by three parameters: $K_d$ for the base color, $K_s$ for the specularity and $g$ for the glossiness.
According to~\citet{sh-rendering}, this light transportation between a Phong BRDF surface and a SH environment map can be efficiently approximated, and we thus employ these rendering equations in our pipeline. 

\paragraph{Hybrid Color Prediction using Transience}
Although the spherical harmonics illumination model typically works well on scenes with ambient environment illumination, it lacks the ability to represent sharp shadows and shiny highlights from high-frequency light sources.
While we believe it is quite impractical to acquire high-frequency details of lighting and material with respect to our unconstrained input, we hope to eliminate the effect caused by those components, and to learn an unbiased result at lower frequencies.
To achieve that, we introduce a hybrid method that combines color prediction with neural networks and parametric models.
As in the geometry network described in Sec.~\ref{sec:method_first_stage}, we employ the concept of \textit{transience}.
However, here we do not learn a separate transient geometry in this model, as our geometry is fixed at this point.
We use the volumetric rendering in Eq.~\ref{eq:NeRF}, but replace the color function with:
\begin{equation}
c_k(\bm{x}) = \text{lerp}\big(c^{(\tau)}_k(\bm{x}), c^{(SH)}(\bm{x}),  \exp(-\sigma^{(\tau)}_k(\bm{x}))\big).
\end{equation}
where $c^{(SH)}(\bm{x})$ is the output color of our SH renderer.

\paragraph{Estimated Depth for Acceleration}
Compared to our geometry networks where color is predicted by neural networks, the rendering stage requires more computation to calculate the color of each sample point due to the more complex rendering equations. On the other hand, however, the learned geometry from the first networks can be used to filter out sampling points that are far away from the object, thus accelerating the whole training process. We develop a hybrid sampling strategy that can speed up the training without introducing any significant artifact. 

For a group of $N_p$ sample points $\bm{x}_i=\bm{r}_o+d_i\bm{r}_d$ on a ray, we build a discrete distribution along the ray with the probability of each point proportional to $\alpha_i(1-w_i)$. Then, we calculate the expectation and variance on $d_i$ w.r.t. to this distribution, denoted as $E(d)$ and $V(d)$. If the variance $V(d)$ is smaller than a threshold $\tau_d$, we then calculate the 3D point at depth $E(d)$ and only use this point for the color calculation. Otherwise, we use all sample points. Please refer to our supplementary material for more details.

\paragraph{Neural Normal Estimation w/ Supervision}
Our network also predicts the final surface normals $\bm{n}(\bm{x})$, supervised by the output of the normal extraction layer in Sec~\ref{sec:method_normal_opt}, with the reconstruction loss $\mathcal{L}_n$ defined by:  
\begin{equation}
\mathcal{L}_n = \left\lVert(||\bm{n}^{(g)}_x||_2) \cdot \bm{n}(x) - \bm{n}^{(g)}_x \right\rVert_2^2,
\end{equation}
We also adopt the normal smoothing loss $\mathcal{L}_{sm}$ in~\citet{DBLP:journals/tog/ZhangSDDFB21} to improve the smoothness of the predicted normals. 

\paragraph{Tone-Mapping}
Since our renderer calculates the radiance in linear HDR space, we also apply a tone-mapping process to the rendered results, defined as:
\begin{equation}
\mathcal{T}_k(x) = x^{(1/\gamma_k)},
\end{equation}
where $\gamma_k$ is a trainable parameter assigned to image $\mathcal{I}_k$, and is initialized from $2.4$, the default value of common sRGB curves. However, we neither apply exposure compensation nor white balance to our renderer's output, assuming that our SH renderer can automatically adapt to these variances during the optimization.

Additionally, to reduce the ambiguity between the material properties and the lighting, we add a regularity loss $\mathcal{L}_\text{reg}$ on both the lighting parameters and material properties. The regularity loss is defined as:
\begin{equation}
\begin{split}
    \mathcal{L}_\text{reg}&=\lambda_\text{spec}\lVert K_s\lVert_2^2 \ + \ \lambda_\text{gamma}\frac{1}{N}\sum_{k=1}^{N}\lVert\gamma_k-2.4\rVert_2^2\\
    &+\lambda_\text{light}\frac{1}{N_t}\sum_{t=1}^{N_t}\lVert\text{ReLU}(-L_{k_t}(\bm{\omega}_t)-\tau_\text{light})\rVert_2^2,
\end{split}
\end{equation}
where $L_{k_t}$ is the environment map of image $k_t$ derived from its SH coefficients, and $\lambda_\text{spec},\lambda_\text{gamma},\lambda_\text{light}$ are coefficients set to $0.1$, $5$, and $5$, respectively. The last term is for light regularization, designed to prevent negative values (lower than $-\tau_\text{light}$, with $\tau_\text{light}$ set to $0.01$) in the SH lighting model, which may happen during training due when overfitting to sharp shadows.
For each iteration, we randomly sample $N_t$ incoming light directions $\bm{\omega}_t$ and image indices $k_t$, and evaluate the corresponding incoming light values for the loss calculation. $N_t$ is set to be identical to the batch size in our experiments.

In summary, the total loss of this stage is defined as:
\begin{equation}
\mathcal{L}_{render} = \mathcal{L}_c + \lambda_{tr}\mathcal{L}_{tr} + \lambda_{n}\mathcal{L}_{n} + \lambda_{sm}\mathcal{L}_{sm} + \mathcal{L}_{reg},
\end{equation}
where the weights $\lambda_{tr}$, $\lambda_{n}$, and $\lambda_{sm}$ are set to $1$, $5$, and $0.5$, respectively, in our experiments.

\subsection{Network Structure}
In our first stage, the geometry network, the input position vector $\bm{x}$ is embedded using the positional encoding method introduced in~\citet{NIPS2017_3f5ee243}, then fed into an 8-layer MLP with a hidden vector dimension of $256$.
The resulting embedding, $\bm{z}_{\bm{x}}$, is then fed into three branches: a branch consisting of one layer to predict static density $\sigma^{(s)}$; a branch consisting of one layer to predict static color $c_k^{(s)}$, which also takes the positional-embedded view direction $\bm{d}$ and appearance embedding $\bm{z}_k^{(a)}$ as input; and a branch of another 4-layer MLP with a hidden vector dimension of $128$, followed by several output layers to predict transient density $\sigma_k^{(\tau)}$, transient color $\bm{c}_k^{(\tau)}$ and uncertainty $\beta_k$, where the transient embedding $\bm{z}_k^{(\tau)}$ is also provided as input.

Our second stage, the rendering network, shares the same structure as the first stage on most components, except for the static color prediction branch.
For this branch, we use a new 4-layer MLP with the hidden vector dimension of $128$, which takes $\bm{x}$ and $\bm{z}_{\bm{x}}$ as input, followed by several output layers to generate the normal $\bm{n}$, base color $K_d$, specularity $K_s$, and glossiness $g$.

We choose ReLU as the activation function for all intermediate layers. For the outputs layers, we adopt SoftPlus for density functions, uncertainty, and glossiness; Sigmoid for static/transient/base color and specularity; and a vector normalizing layer for normal estimation.

In addition to our network parameters, we also jointly optimize the light coefficients $L_{k,lm}$, the camera parameters $(\delta_R,\delta_t,\delta_f)_k$, and the tone-mapping parameter $\gamma_k$ for each image $\mathcal{I}_k$.

\section{Evaluations}
\label{sec:eval}

\subsection{Implementation details}

\paragraph{Training}
We use a modified version of MLP structure following \citet{mildenhall2020nerf,DBLP:conf/cvpr/Martin-BruallaR21} as our networks. In the training, We use the Adam optimizer~\cite{DBLP:journals/corr/KingmaB14} to learn all of our parameters, and our initial learning rate is set to $4\cdot 10^{-4}$.
Our training and inference experiments are implemented using the PyTorch framework~\cite{NEURIPS2019_9015}.
We train our model on 4 NVIDIA V100s with the batch size of 4096, and test our model on a single NVIDIA V100. In the first stage, we train our model with 30 epochs (60K-220K iterations), in roughly 6 to 13 hours. We then apply our normal extraction layer to generate the surface normals for all training images within 1 hour. For the second stage, we freeze the static MLP for density prediction and train the rest of our model from scratch. Approximately 2 to 4 hours are required for 10 epochs of training. 

\paragraph{Datasets}
We use datasets of 13 objects in our evaluations, collected from three different sources: image collections found on the Internet; objects we captured ourselves; and data published in NeRD~\cite{DBLP:conf/iccv/BossBJBLL21}. The number of training images of each object varies from 40 to 200. For more details on our datasets, please refer to the supplementary material. 


\subsection{Comparisons}

\definecolor{bestcolor}{rgb}{1, 0.5, 0.25}
\definecolor{secondbestcolor}{rgb}{1, 0.8, 0.5}
\newcommand{\bone}{\cellcolor{bestcolor}}
\newcommand{\btwo}{\cellcolor{secondbestcolor}}

\begin{figure}[h]
\begin{center}
\newcommand{\img}[1]{\includegraphics[height=0.12\textwidth]{#1}}
\newcommand{\imgl}[1]{\includegraphics[width=0.1\textwidth]{#1}}
\begin{tabular}{cccc}

\footnotesize GT & \footnotesize NERF & \footnotesize Ours-Geom & \footnotesize Ours-Full \\

\imgl{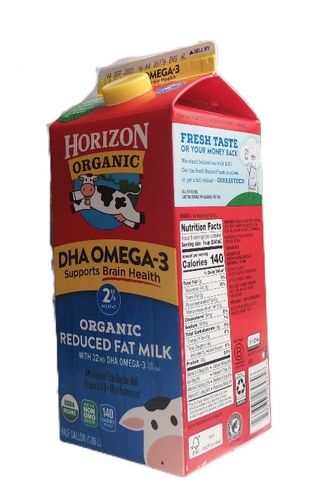} & 
 \imgl{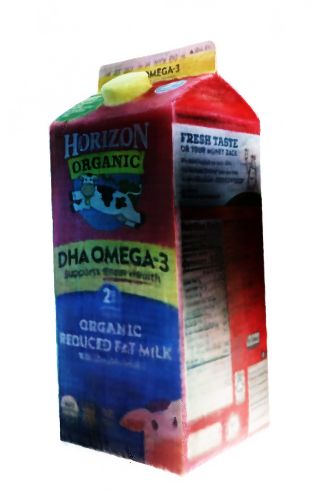} & 
\imgl{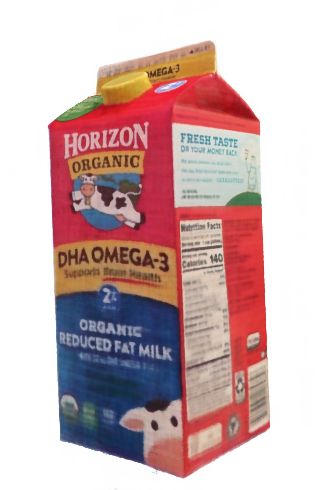} & 
 \imgl{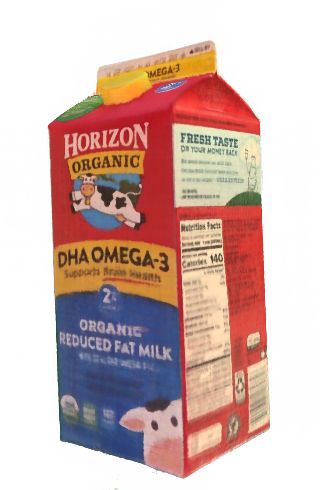} \\

\imgl{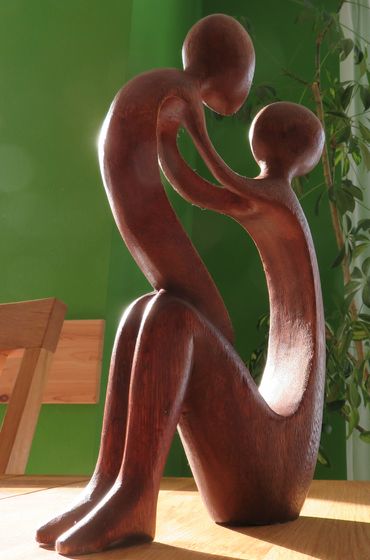} & 
\imgl{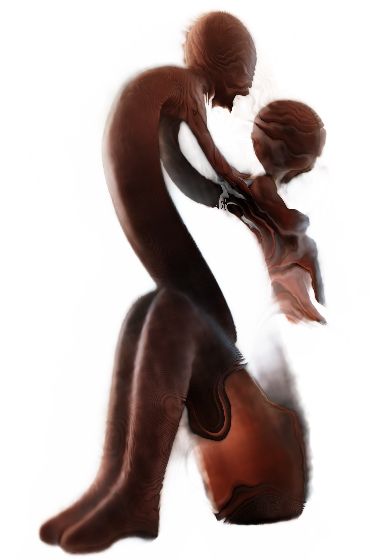} & 
\imgl{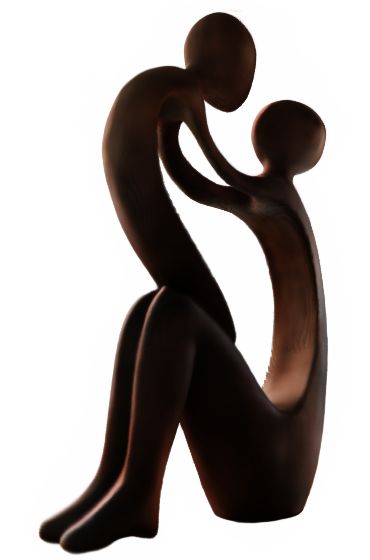} & 
\imgl{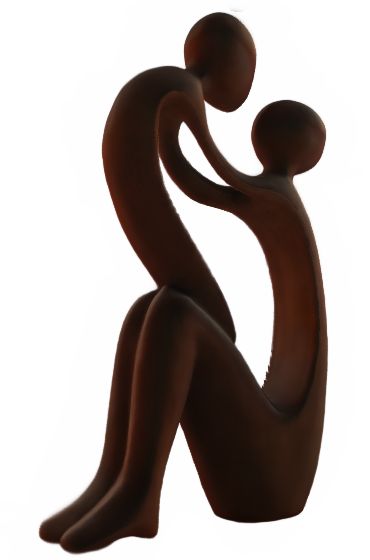} \\

\imgl{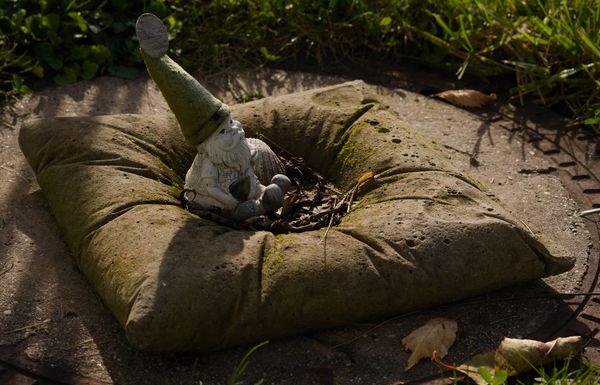} &
 \imgl{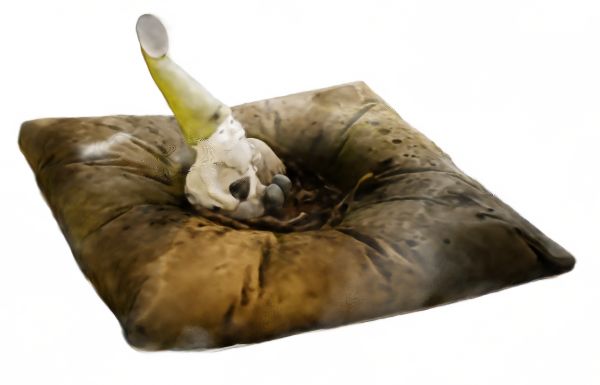} &
\imgl{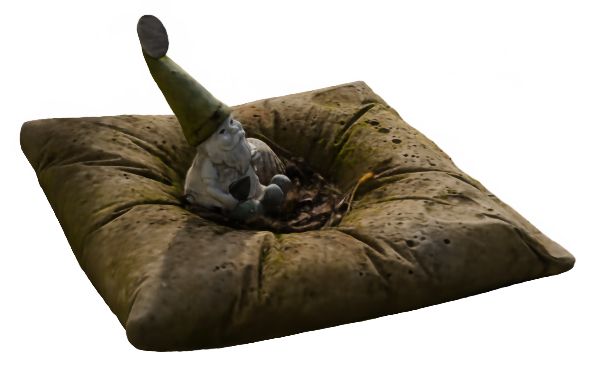} &
 \imgl{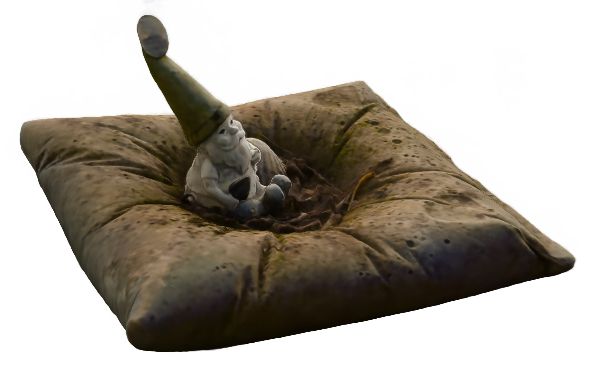}
\end{tabular}
\end{center}
\caption{
\textbf{Qualitative Comparisons with NeRF.} Result of \textit{Ours-Full} are rendered with SH rendering only. Some shadows and highlights are handled as transient component, thus not appearing in \textit{Ours-Full}.}
\label{fig:qual_compare_nerf}

\end{figure}

We first show comparisons between our model and NeRF~\cite{mildenhall2020nerf} in 7 offline captured objects (Milk, Figure, TV, Gnome, Head, Cape, MotherChild). We adopt the commonly used metrics Peak Signal-to-Noise Ratio (PSNR), Structural Similarity Index Measure (SSIM), and Learned Perceptual Image Patch Similarity (LPIPS) in our evaluation. Considering that the illumination conditions of the testing images are unknown, we propose two settings to evaluate our model: using the lighting parameters (e.g. the embedded vector for the geometry network, and the SH coefficients and the gamma value for the rendering network) from another training image in the same scene; or freezing the networks and optimize the lighting parameters with a Stochastic Gradient Descent optimizer for 1000 steps. As shown in Tab.~\ref{tab:compare_NeRF} and Fig.~\ref{fig:qual_compare_nerf}, our method outperforms NeRF in both settings by a considerable margin, except on the NeRD Head dataset. We believe that this is because it is a relatively simple case in which the object is rotating in front of a camera with a fixed pose. Since NeRF has fewer components and parameters to optimize, it is more likely to converge to a sharper geometry and get better testing results in such a trivial scenario. 
In qualitative results, our model generates more consistent and smooth results than NeRF does. Aside from the comparison with NeRF, we would like to highlight that our rendering networks produce competitive results compared to the first stage, while it also supports relighting on unseen environments.

We also compare our model with NeRF on the TV dataset with fewer training images. In this experiment, we train our model and NeRF on three training sets consisting of all images, 20 images and 10 images respectively. We then test them on the same testing set. The results are shown in Tab.~\ref{tab:comparison_imgnum}. As scores of both model are both decreased when training images get fewer, our model are less affected than NeRF, especially on LPIPS score. This result proves that our model is able to keep the output accurate in a perceptual manner even if only sparse training images are given.

\begin{table*}[t]
\caption{\textbf{Comparison with NeRF~\cite{mildenhall2020nerf}}. We report the PSNR, SSIM and LPIPS scores of 7 tested scenes. We also ablate on w/ only the first stage of our approach (Ours-Geom) and w/o appearance optimization (w/o Opt). We highlight the best and second best results of each column in orange and yellow, respectively. The last column contains the mean result across all target objects.
  }
\begin{center}
\resizebox{0.82\linewidth}{!}{
{\small
\begin{tabular}{|c| c|c|c| c|c|c| c|c|c| c|c|c|}
\hline
\multirow{2}{*}{Methods} & \multicolumn{3}{c|}{Cape} & \multicolumn{3}{c|}{Gnome} & \multicolumn{3}{c|}{Head} & \multicolumn{3}{c|}{MotherChild}\\

& PSNR$\uparrow$ & SSIM$\uparrow$ & LPIPS$\downarrow$  & PSNR$\uparrow$ & SSIM$\uparrow$ & LPIPS$\downarrow$  & PSNR$\uparrow$ & SSIM$\uparrow$ & LPIPS$\downarrow$  & PSNR$\uparrow$ & SSIM$\uparrow$ & LPIPS$\downarrow$ \\
 
\hline 

NeRF~\shortcite{mildenhall2020nerf}
        & 22.95         & 0.78          & 0.218         & 18.73         & 0.82          & 0.240         & \bone 29.13   & \bone 0.92    & 0.140      
        & 25.08         & 0.95          & 0.106         \\

Ours-Geom w/o Opt
        & 23.22         & \btwo 0.82    & \btwo 0.180   & \btwo 26.95   & \bone 0.89    & \btwo 0.120   & 26.37         & \bone 0.92    & \bone 0.136 
        & 23.03         & 0.95          & 0.068         \\

Ours-Full w/o Opt   
        & 22.82         & 0.79          & 0.198         & 25.30         & 0.87          & 0.132         & 26.08         & 0.90          & 0.146      
        & 25.69         & 0.96          & 0.069         \\

Ours-Geom 
        & \bone 24.70   & \bone 0.83    & \bone 0.178   & \bone 28.11   & \bone 0.89    & \btwo 0.119   & \btwo 26.80   & \bone 0.92    & \bone 0.136 
        & \btwo 28.98   &\bone 0.97     & \bone 0.058   \\

Ours-Full 
        & \btwo 23.47   & 0.80          & 0.197         & 26.11         & 0.88         & 0.129          & 26.35         & 0.91          & 0.145     
        & \bone 29.02   &\bone 0.97     & \btwo 0.062   \\

\hline

\multirow{2}{*}{Methods} & \multicolumn{3}{c|}{Milk} & \multicolumn{3}{c|}{Figure} & \multicolumn{3}{c|}{TV} & \multicolumn{3}{c|}{\textbf{Mean}} \\

& PSNR$\uparrow$ & SSIM$\uparrow$ & LPIPS$\downarrow$  & PSNR$\uparrow$ & SSIM$\uparrow$ & LPIPS$\downarrow$  & PSNR$\uparrow$ & SSIM$\uparrow$ & LPIPS$\downarrow$  & PSNR$\uparrow$ & SSIM$\uparrow$ & LPIPS$\downarrow$ \\

\hline

NeRF~\shortcite{mildenhall2020nerf}
        & 19.40         & 0.92          & 0.145         & 21.54         & 0.90          & 0.159         & 19.03         & 0.90          & 0.145 
        & 22.266        & 0.884         & 0.164         \\
Ours-Geom w/o Opt
        & 21.41         & 0.94          & 0.059         & 22.89         & \btwo 0.92    & \btwo 0.121   & 20.31         & 0.92          & 0.114 
        & 23.454        & \btwo 0.909   & \btwo 0.114   \\
Ours-Full w/o Opt   
        & 23.00         & \btwo 0.95    & 0.066         & 23.69         & \btwo 0.92    & 0.133         & 21.88         & 0.92          & 0.122 
        & 24.065        & 0.901         & 0.124         \\
Ours-Geom 
        & \btwo 27.51   & \bone 0.96    & \bone 0.052   & \btwo 24.41   & \bone 0.93    & \bone 0.118   & \btwo 25.06   & \bone 0.93    & \btwo 0.110
        & \bone 26.510  & \bone 0.919   & \bone 0.110   \\
Ours-Full 
        & \bone 28.87   & \btwo 0.95    & \btwo 0.056   & \bone 24.72   & \btwo 0.92    & 0.130         & \bone 26.52   & \bone 0.93    & \bone 0.107
        & \btwo 26.437  & \btwo 0.909   & 0.118         \\

\hline

\end{tabular}}}
\end{center}
\label{tab:compare_NeRF}
\end{table*}

\setlength{\tabcolsep}{1.5pt}

\begin{table*}[t]
\caption{\textbf{Comparison with NeRF with fewer images}. We highlight the best and second best results in each column in orange and yellow.}
\begin{center}
\newcolumntype{P}[1]{>{\centering\arraybackslash}p{#1}}

{\small
\begin{tabular}{|P{0.13\linewidth}|P{0.068\linewidth}|P{0.068\linewidth}|P{0.068\linewidth}|P{0.068\linewidth}|P{0.068\linewidth}|P{0.068\linewidth}|P{0.068\linewidth}|P{0.068\linewidth}|P{0.068\linewidth}|}
\hline
\multirow{2}{*}{Methods} & \multicolumn{3}{c|}{Full} & \multicolumn{3}{c|}{w/ 20 Images} & \multicolumn{3}{c|}{w/ 10 Images}  \\

& PSNR$\uparrow$ & SSIM$\uparrow$ & LPIPS$\downarrow$  & PSNR$\uparrow$ & SSIM$\uparrow$ & LPIPS$\downarrow$  & PSNR$\uparrow$ & SSIM$\uparrow$ & LPIPS$\downarrow$ \\

\hline

NeRF~\shortcite{mildenhall2020nerf}
& 19.03 & 0.90 & 0.145
& 16.52 & 0.89 & 0.167
& 17.51 & 0.88 & 0.244 \\
Ours-Geom 
& \btwo 25.06 & \bone 0.93 & \btwo 0.110 
& \btwo 24.76 & \bone 0.92 & \bone 0.119 
& \btwo 23.54 & \btwo 0.91 & \btwo 0.148 \\
Ours-Full 
& \bone 26.52 & \bone 0.93 & \bone 0.107
& \bone 26.26 & \bone 0.92 & \bone 0.119
& \bone 25.26 & \bone 0.92 & \bone 0.147 \\
\hline
\end{tabular}}
\end{center}
\label{tab:comparison_imgnum}
\end{table*}

\begin{figure}[t]
\begin{center}
\includegraphics[width=\columnwidth]{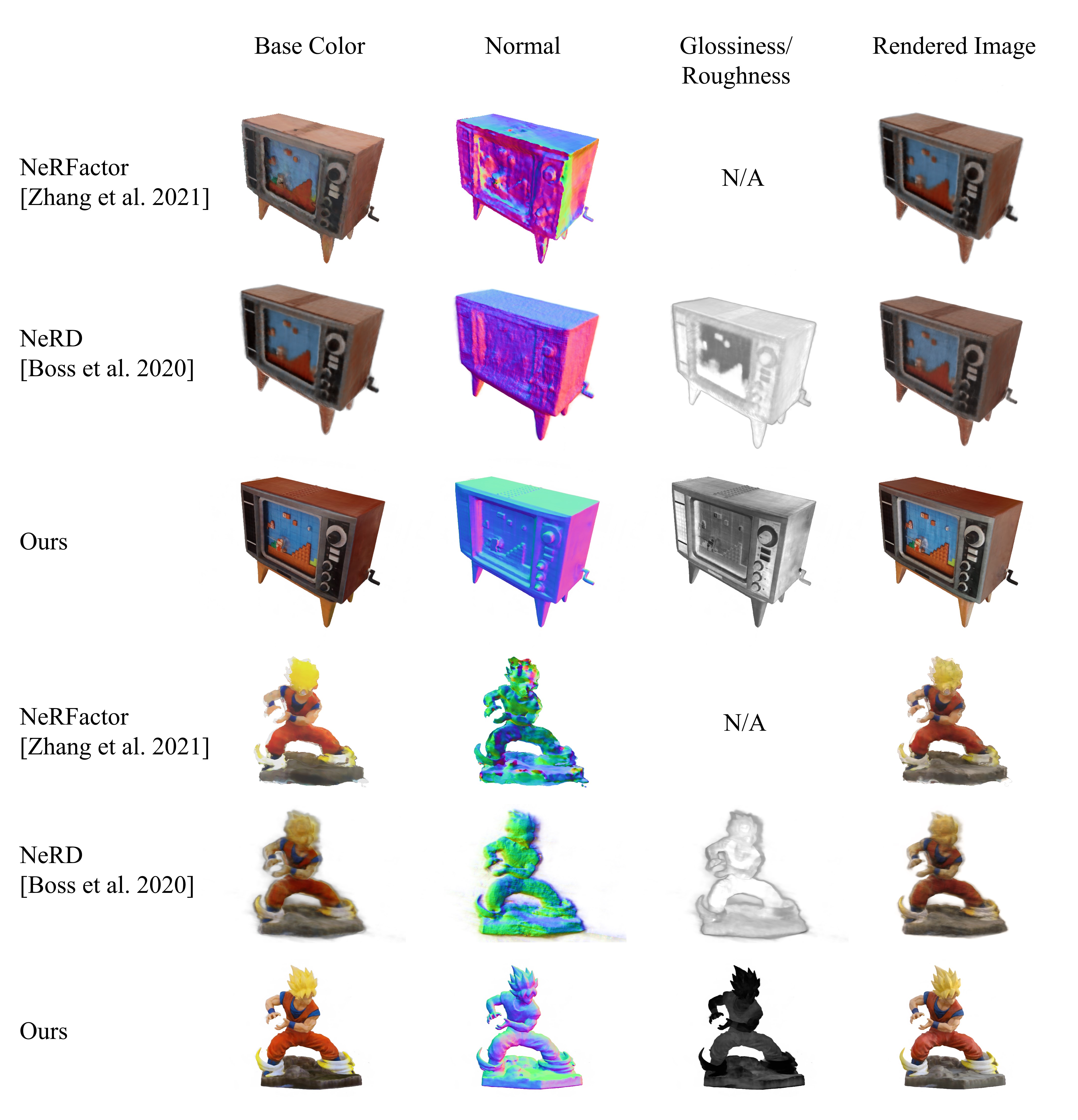}
\end{center}
\caption{
\textbf{Comparison on Decomposition.} We show diffuse albedo, surface normal, roughness/glossiness maps and final rendered result of the TV and Figure datasets, predicted by NeRFactor~\cite{DBLP:journals/tog/ZhangSDDFB21}, NeRD~\cite{DBLP:conf/iccv/BossBJBLL21} and our method. Note that the normal and roughness/glossiness maps of each method are encoded in different ways from respective code bases, and NeRFactor does not explicitly predict a roughness map. We also note that NeRFactor was originally designed for datasets with constant lighting environments.}
\label{fig:qual_compare_material}
\end{figure}

\begin{figure}[h]
\begin{center}
\newcommand{\img}[1]{\includegraphics[height=0.32\linewidth]{#1}}
\setlength{\tabcolsep}{1pt}
\begin{tabular}{ccc}
\img{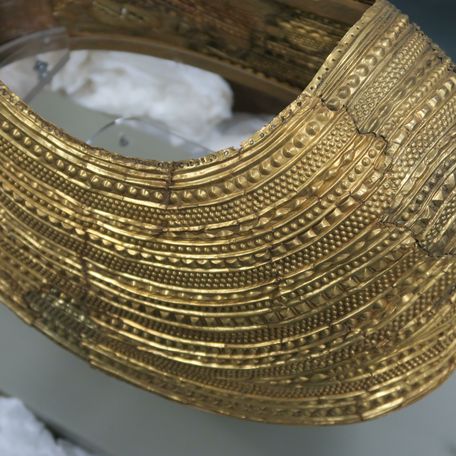} & 
\img{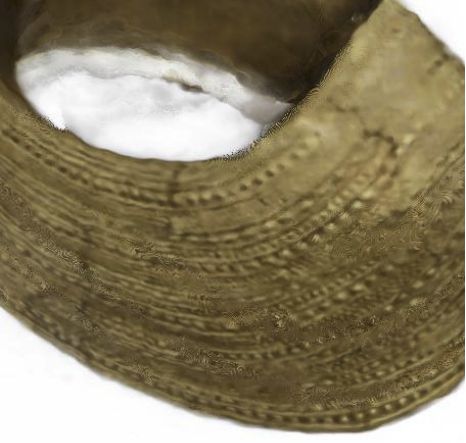} & 
\img{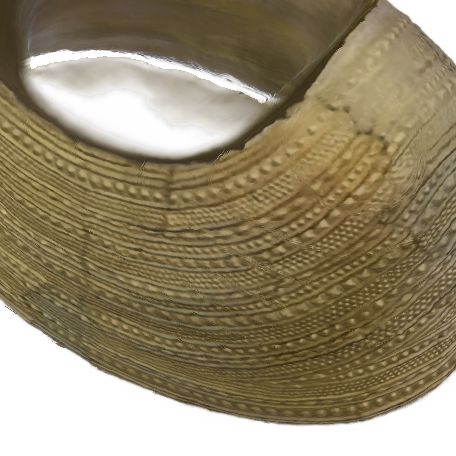} \\

\img{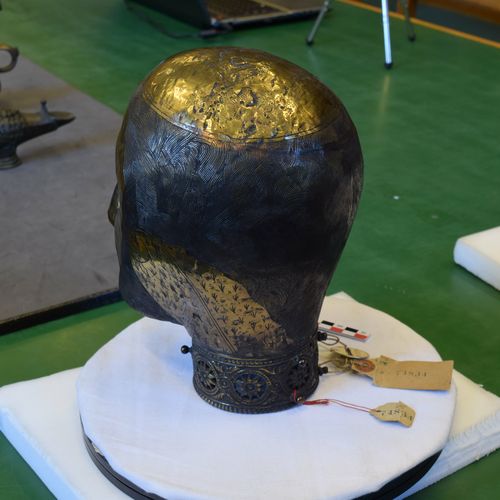} & 
\img{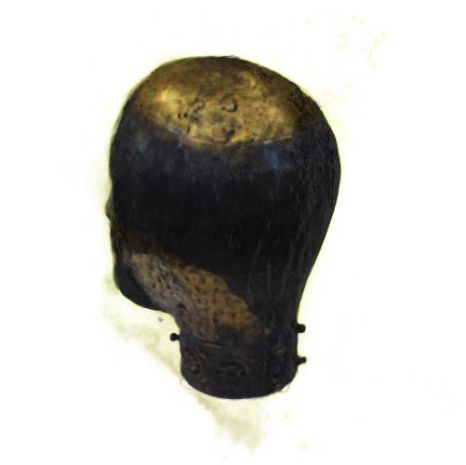} & 
\img{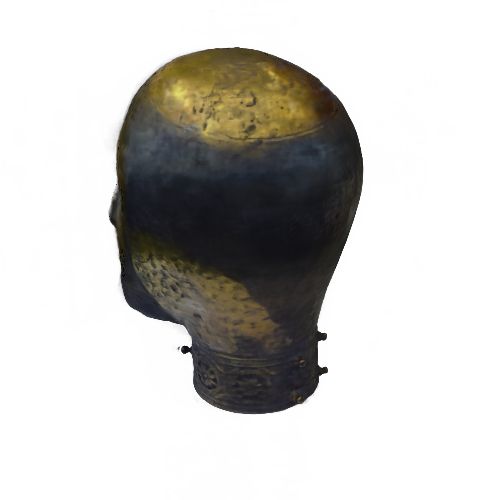} \\

\footnotesize Ground Truth & \footnotesize NeRD~\cite{DBLP:conf/iccv/BossBJBLL21} & \footnotesize Ours 
\end{tabular}
\end{center}
\caption{
\textbf{Comparison with NeRD on their data}. Results from NeRD are copied from the paper. Thanks to the novel designs of our model, our model achieves better results with smoother geometry, cleaner object boundaries, and sharper textures compared to results shown in NeRD.}
\label{fig:qual_compare_nerd}
\end{figure}

Aside from NeRF, we compare two state-of-the-art decomposition methods with our approach: NeRD~\cite{DBLP:conf/iccv/BossBJBLL21} and NerFactor~\cite{DBLP:journals/tog/ZhangSDDFB21}. Results are shown in Fig.\ref{fig:qual_compare_nerd} and Fig.\ref{fig:qual_compare_material}. We found that our approach performed well compared to NeRD when using their data, and outperforms both NeRD and NerFactor on the less constrained images we use, which contain a wider range of backgrounds and lighting conditions. For more comparisons and discussions of these works, please consult the supplementary material.

\subsection{Ablations}

To help understand the importance and effectiveness of our contributions, we further conduct four ablative studies on our model.

\paragraph{Ablation Study on Novel View Synthesis}
In the first study, we quantitatively compare our geometry and rendering model with variants on the MotherChild dataset. For our geometry model, the variants we evaluated are: Our model trained without silhouette loss (\textit{Model w/o sil}); without adaptive sampling (\textit{Model w/o ada}); and without transient components (\textit{Model w/o tr}). Variants of our rendering model are: Our model trained without regularity loss (\textit{Model w/o reg}); without transient components (\textit{Model w/o tr}); and without the normal extraction layer (\textit{Model w/o NEL}). We report the PSNR score for all models, and the mean squared error between the attenuation map and the grount-truth foreground mask (denoted as MMSE) for all geometry models. The results are shown in Tab.~\ref{tab:ablation_1}. Our model generally outperforms all variants on both stages, except in two cases. First, removing the transient component from the geometry model did not significantly affect its performance on the PSNR metric. We believe this is because, in most images, the areas with occlusions and incorrect masking are relatively small compared to the object, and thus have a minor effect on the color reconstruction. However, it still decreases the accuracy of the geometry silhouette by one order of magnitude. Second, removing the normal extraction layer from the rendering model did not notably decrease its optimization-free PSNR score either. We believe this is because the effect on the normal is canceled out by other components when rendering the testing view. We will directly compare the normal maps in our next experiment.

\begin{table}[h]
\begin{center}
\caption{\textbf{Ablation Study on Novel View Synthesis}. We show PSNR and MMSE scores on our geometry network models (top 4 rows) and rendering network models (bottom 4 rows). 
For the geometry network, from top to bottom: model trained without silhouette loss; model trained without adaptive sampling; model trained without transient model; full model. 
For rendering network, from top to bottom: model trained without regularity loss; model trained without transient model; model trained without normal extraction layer; full model. Please notice that MMSE is a geometry-based metric and thus not affected by model optimization nor rendering network training.}

\label{tab:ablation_1}
\newcolumntype{P}[1]{>{\centering\arraybackslash}p{#1}}
{\small
\begin{tabular}{P{40pt}|c|c|P{40pt}|P{40pt}}
Stage & Methods & PSNR w/o opt$\uparrow$ & PSNR$\uparrow$ & MMSE$\downarrow$ \\
\hline
\multirow{4}{*}{Geometry} & Model w/o sil &
21.30&23.38&0.180\\
& Model w/o ada &
13.60&13.78&0.073\\
& Model w/o tr &
\btwo 21.34& \bone 30.27& \btwo 0.030\\
& Full model&
\bone 23.03& \btwo 28.98& \bone 0.003 \\
\hline\hline
\multirow{4}{*}{Rendering} & Model w/o reg &
25.57 & 28.00 & \multirow{4}{*}{N/A} \\
 & Model w/o tr &
24.95 & 28.03 & \\
 & Model w/o NEL &
\bone 25.79 & \btwo 28.53 & \\
 & Full model &
\btwo25.69 & \bone 29.02 & \\

\end{tabular}}
\end{center}
\end{table}

\begin{figure}[h]
\begin{center}
\newcommand{\img}[1]{\includegraphics[width=0.055\textwidth]{#1}}
\begin{tabular}{c}

\includegraphics[width=\columnwidth]{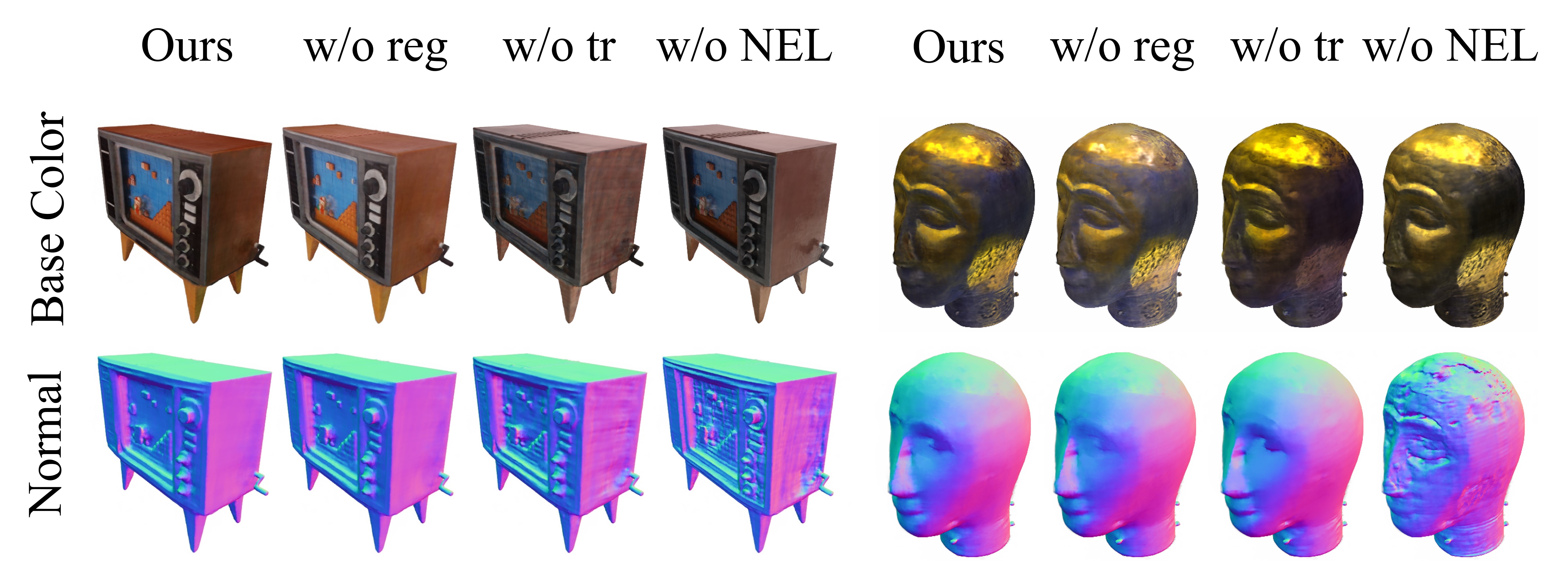}

\end{tabular}
\end{center}
\caption{
\textbf{Qualitative Results of Ablation Study on Rendering Network.} We show diffuse albedo maps and normal maps predicted by our models. We increase the exposure of albedo maps for the Head data since the black area in the original outputs is extremely dim.
}
\label{fig:qual_ablation}
\end{figure}

\paragraph{Ablation Study on Decomposition}
Our second ablation study aims to demonstrate that our specific approach generates smooth and accurate material properties in the second stage. We qualitatively compare our full model with three variants mentioned above on the TV and Head datasets. As shown in Fig.~\ref{fig:qual_ablation}, using the normals without proper processing results in worse normal predictions for both datasets; removing the transient component leads to unwanted artifacts on the TV's smooth surface, where mirroring reflections are more likely to appear; and removing the regularity loss results in a biased albedo for the Head data. Our full model addresses each of these problems, producing the most appropriate results.

\begin{figure}[t]
\begin{center}
\includegraphics[width=\columnwidth]{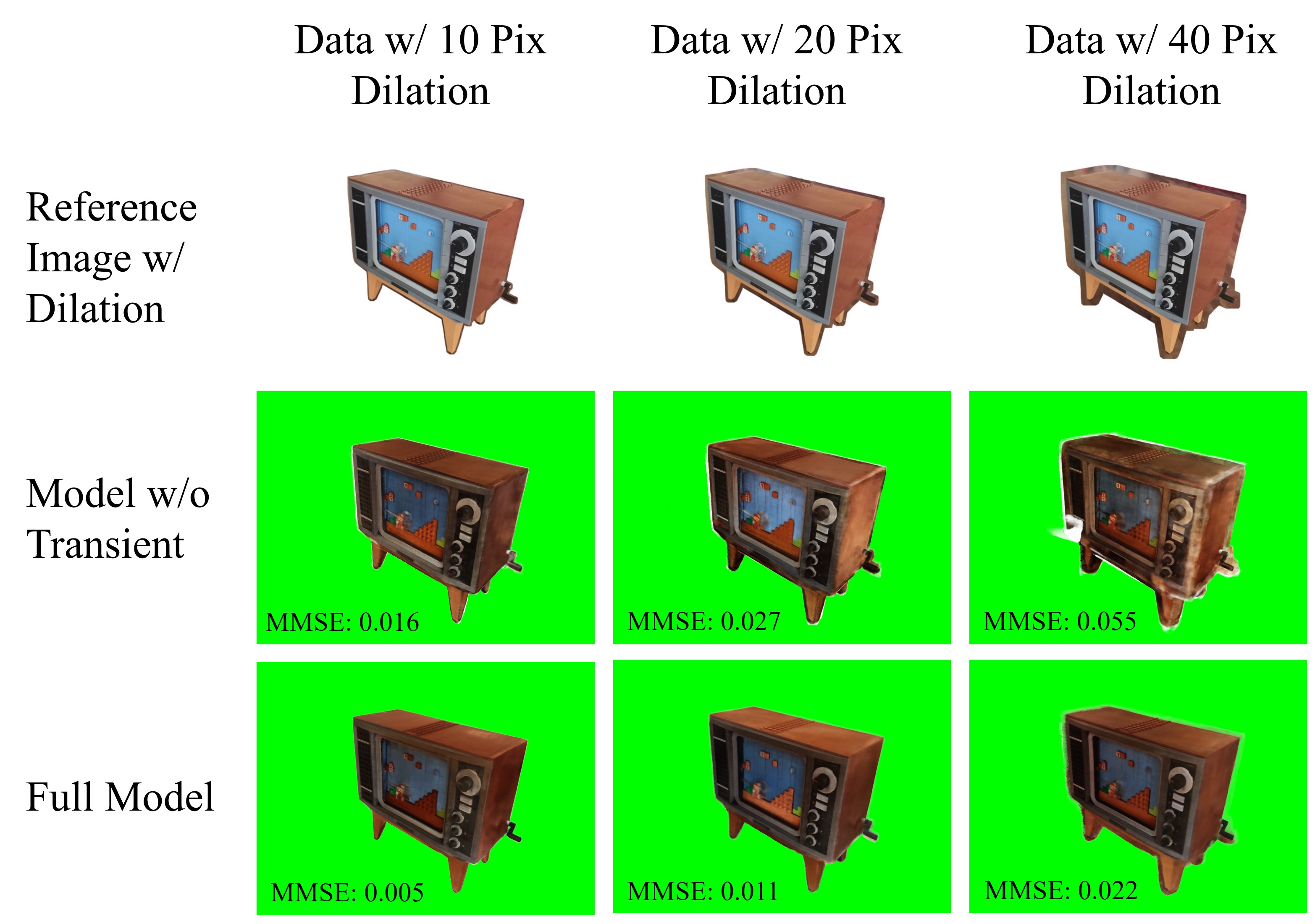}
\end{center}
\caption{
\textbf{Ablation Study on Imperfect Masks.} Results of our full model and model without transient embedding (Model w/o Transient) on the TV dataset with imperfect masks. The average MMSE scores on all testing images are also shown at the left-bottom corner of each result image.}
\label{fig:qual_imperfect_mask}
\end{figure}

\paragraph{Ablation Study on Imperfect Masks}
While our model performs well on our data with pre-processed foreground masks, we also evaluate our approach's capacity to handle improperly segmented input images, testing our model in an alternative, more challenging setting with imperfect masks. In this experiment, we dilate the foreground masks of all training images of TV data by multiple fixed radii (10, 20 and 40 pixels), and evaluate our full geometry model and our model trained without transient components. This dilation process aims to simulate the common error of roughly annotated masks, where the mask contains the object but does not fit the object silhouette perfectly. We also emphasize that this setting is more challenging than many real cases. As all training images share an identical type of mask dilation, it becomes harder for the model to distinguish the foreground object from the wrongly annotated background. The results are shown in Fig.\ref{fig:qual_imperfect_mask}. We note that our model generally reconstructs the geometry of the TV well with mask dilations within 40 pixels, and removing the transient component significantly affects the robustness of our model.

\paragraph{Ablation Study on Camera Optimization}
Our last ablation study focuses on evaluating the effect of the camera optimization technique in our geometry model in improving its robustness to pose perturbations. We use the synthetic datasets from NeRD (Chair, Car and Globe), which contain ground-truth poses. For each dataset, we only use 40 images of its training set to train our model. Three types of perturbations are added to the poses: (1) rotational perturbation within a certain degree range ($10^{\circ} - 20^{\circ}$); (2) translational perturbation within a certain distance ($0.1 - 0.2$, after the poses are normalized); and (3) the combination of the first two perturbations. We test our full model and our model trained without camera optimization (Model w/o cam-opt), and report the average PSNR scores, the number of scenes where each model succeed to converge (Succ No.), and an error defined by the average mean squared error between the ground truth fundamental matrices and the predicted ones (FMSE). Specifically, this error is defined as:
\begin{equation}
    \mathcal{E}_\text{FMSE} = \frac{1}{N_t(N_t-1)} \sum_{i,j\in N_t} \lVert F^{gt}_{i,j} - F^{pred}_{i,j} \rVert_2
\end{equation}
where $F_{i,j}$ is the fundamental matrix derived from the poses of the i-th and j-th images. This error is to measure the accuracy of the camera poses without the affect of global transformations of the scene or changes to the camera intrinsic parameters, which can occur during a joint optimization of the model and the camera poses.

The results are shown in Tab.\ref{tab:ablation_camera}.
Our model successfully converged for all scenes when the rotational perturbation is under $10^{\circ}$ (rotational) and the translational perturbation is under 0.1. Also, removing the camera optimization from our model hugely affects the robustness of our model when handling inaccurate camera poses. 

\begin{table*}[ht]
\caption{\textbf{Ablation study on camera optimization}. We highlight the best results in each column in orange. We also would like to mention that our full model converged on all scenes where the model w/o cam-opt converged as well. For PSNR metric, we report the average score of all scenes where our model converged.}
\begin{center}
\newcolumntype{P}[1]{>{\centering\arraybackslash}p{#1}}

{\small
\begin{tabular}{|P{0.13\linewidth}|P{0.068\linewidth}|P{0.068\linewidth}|P{0.068\linewidth}|P{0.068\linewidth}|P{0.068\linewidth}|P{0.068\linewidth}|P{0.068\linewidth}|P{0.068\linewidth}|P{0.068\linewidth}|}
\hline

\multirow{2}{*}{Methods} 
& \multicolumn{3}{c|}{$\Delta t=0.1$}
& \multicolumn{3}{c|}{$\Delta t=0.2$ }
& \multicolumn{3}{c|}{$\Delta r=10^{\circ}$} \\

& PSNR$\uparrow$ & FMSE$\downarrow$ & Succ No.$\uparrow$  & PSNR$\uparrow$ & FMSE$\downarrow$ & Succ No.$\uparrow$  & PSNR$\uparrow$ & FMSE$\downarrow$ & Succ No.$\uparrow$ \\

\hline

Model w/o cam-opt & 21.20 & 0.00031 & \bone 3/3 & 17.82 & 0.00124 & 1/3 & 19.36 & 0.00264 & \bone 3/3 \\
Full Model & \bone 24.74 & \bone 0.00029 & \bone 3/3 & \bone 23.81 & \bone 0.00054 & \bone 2/3 & \bone 22.67 & \bone 0.00146 & \bone 3/3 \\

\hline

& \multicolumn{3}{c|}{$\Delta r=20^{\circ}$}
& \multicolumn{3}{c|}{$\Delta r=10^{\circ}, \Delta t=0.1$ }
& \multicolumn{3}{c|}{$\Delta r=20^{\circ}, \Delta t=0.2$ } \\

& PSNR$\uparrow$ & FMSE$\downarrow$ & Succ No.$\uparrow$  & PSNR$\uparrow$ & FMSE$\downarrow$ & Succ No.$\uparrow$  & PSNR$\uparrow$ & FMSE$\downarrow$ & Succ No.$\uparrow$ \\

\hline

Model w/o cam-opt & 18.61 & 0.01048 & \bone 2/3 & 20.18 & 0.00300 & \bone 3/3 & 15.94 & 0.01190 & \bone 2/3 \\
Full Model & \bone 20.38 & \bone 0.00860 & \bone 2/3 & \bone 22.22 & \bone 0.00150 & \bone 3/3 & \bone 20.28 & \bone 0.00945 & \bone 2/3 \\

\hline

\end{tabular}}
\end{center}
\label{tab:ablation_camera}
\end{table*}

\subsection{Relighting and Compositing Results}
We provide more results of our model in three different showcases: material decomposition, relighting, and composition with online objects. 

\begin{figure}[h]
\begin{center}
\newcommand{\img}[1]{\includegraphics[width=0.11\textwidth]{#1}}
\setlength{\tabcolsep}{1.0pt}
\begin{tabular}{cccc}
\img{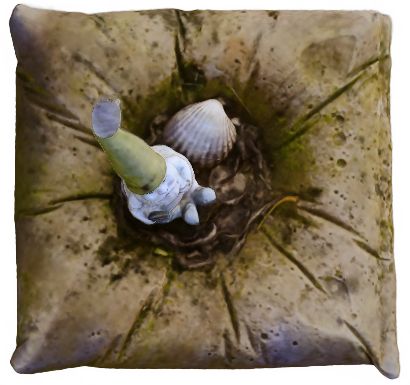} & 
\img{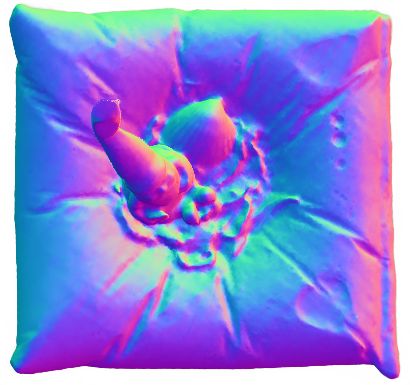} & 
\img{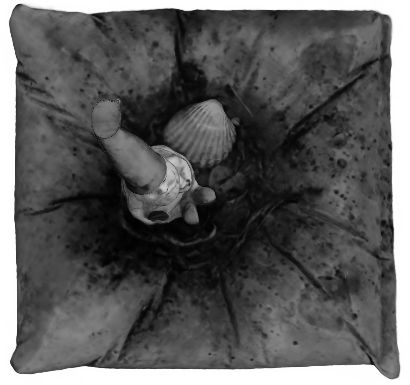} & 
\img{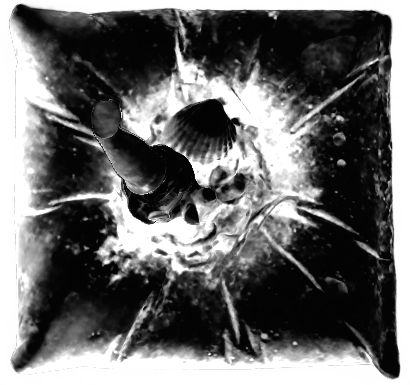} \\


\img{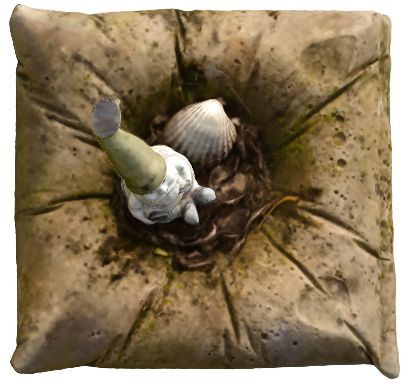} & 
\img{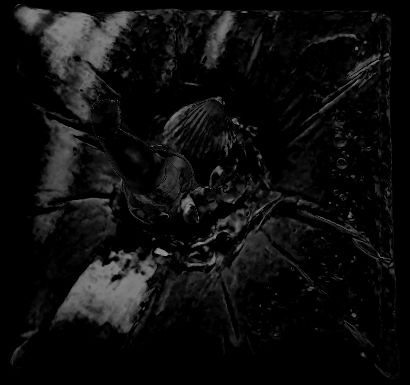} & 
\img{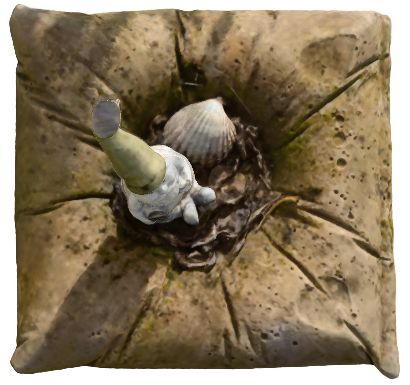} & 
\img{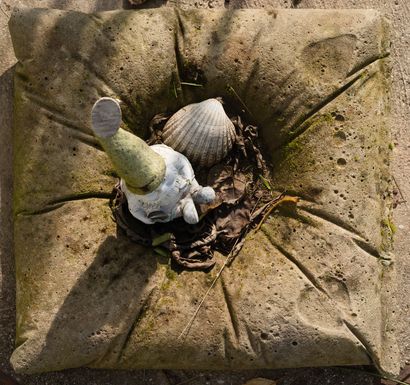} \\
\end{tabular}
\end{center}
\caption{
\textbf{Examples of Our Decomposition Results.}
\textit{First row}: from left to right: diffuse albedo map, normal map, specularity map, and Glossiness map. \textit{Second row}: from left to right: color rendering w/o transient, transient blending weight, color rendering w/ transient, and ground-truth image.}
\label{fig:material_decomp}
\end{figure}

\begin{figure}[t]
\begin{center}
\newcommand{\img}[1]{\includegraphics[width=0.11\textwidth]{#1}}
\setlength{\tabcolsep}{2.0pt}
\begin{tabular}{cccc}

\multicolumn{2}{c|}{\footnotesize Pose \#1} & \multicolumn{2}{c}{\footnotesize Pose \#2} \\
\hline 
\footnotesize Real & \footnotesize Light \#1 & \footnotesize Light \#2 & \footnotesize Light \#3 \\

\img{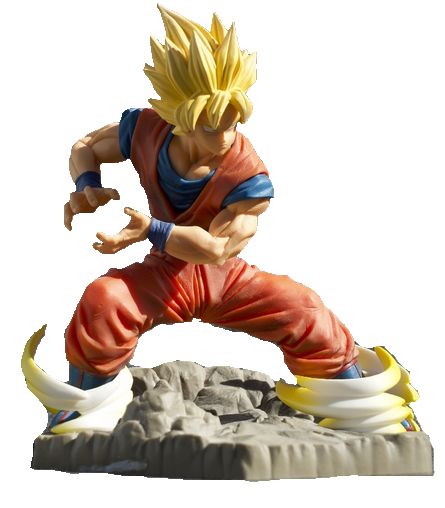} & 
\img{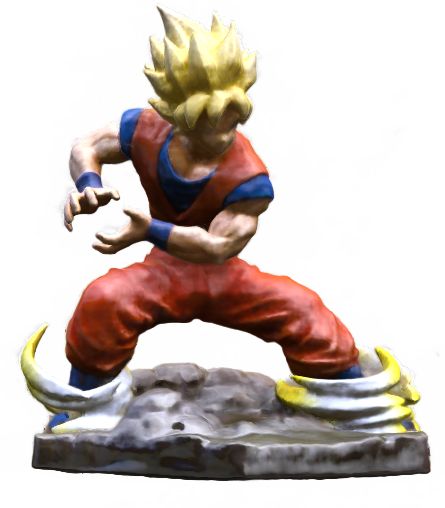} & 
\img{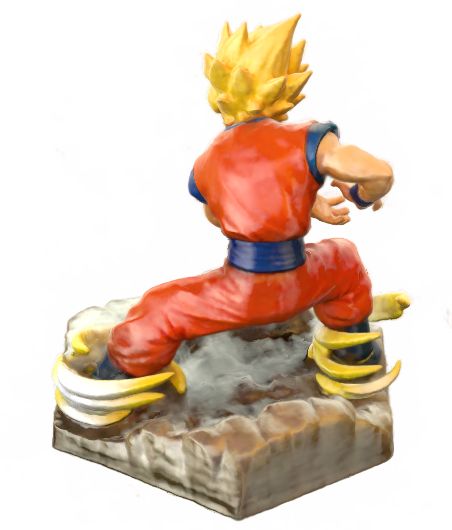} & 
\img{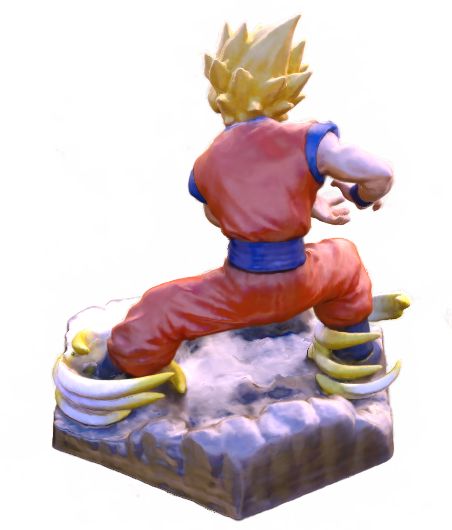} \\

\img{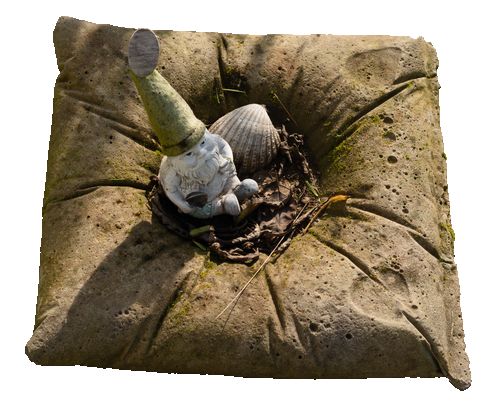} & 
\img{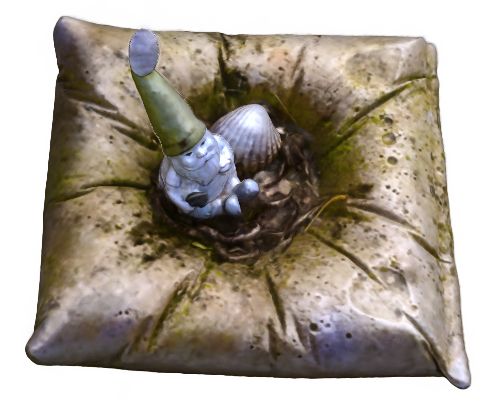} & 
\img{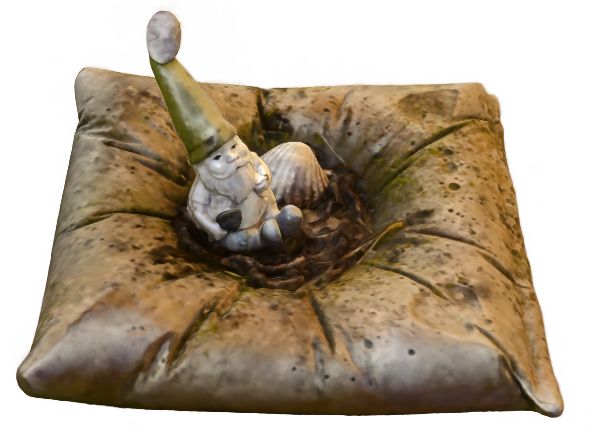} & 
\img{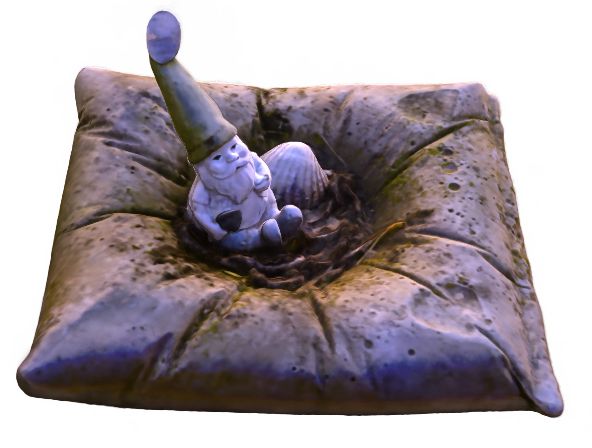} \\

\end{tabular}
\end{center}
\caption{
\textbf{Relighting Results.} Left: comparison between our relit result and a real image; Right: Model relit in another pose.}
\label{fig:relight_results}
\end{figure}

Fig.~\ref{fig:material_decomp} shows our prediction of material properties and rendering components for the Gnome dataset. We would like to highlight that our model successfully disentangled the shadow in the input images from our SH rendering component, and learned unbiased material properties from the whole training set. With the material properties, we are able to re-render the objects with new lighting environments, and the results are shown in Fig.~\ref{fig:relight_results}. 

\begin{figure}[t]
\begin{center}
\includegraphics[width=\columnwidth]{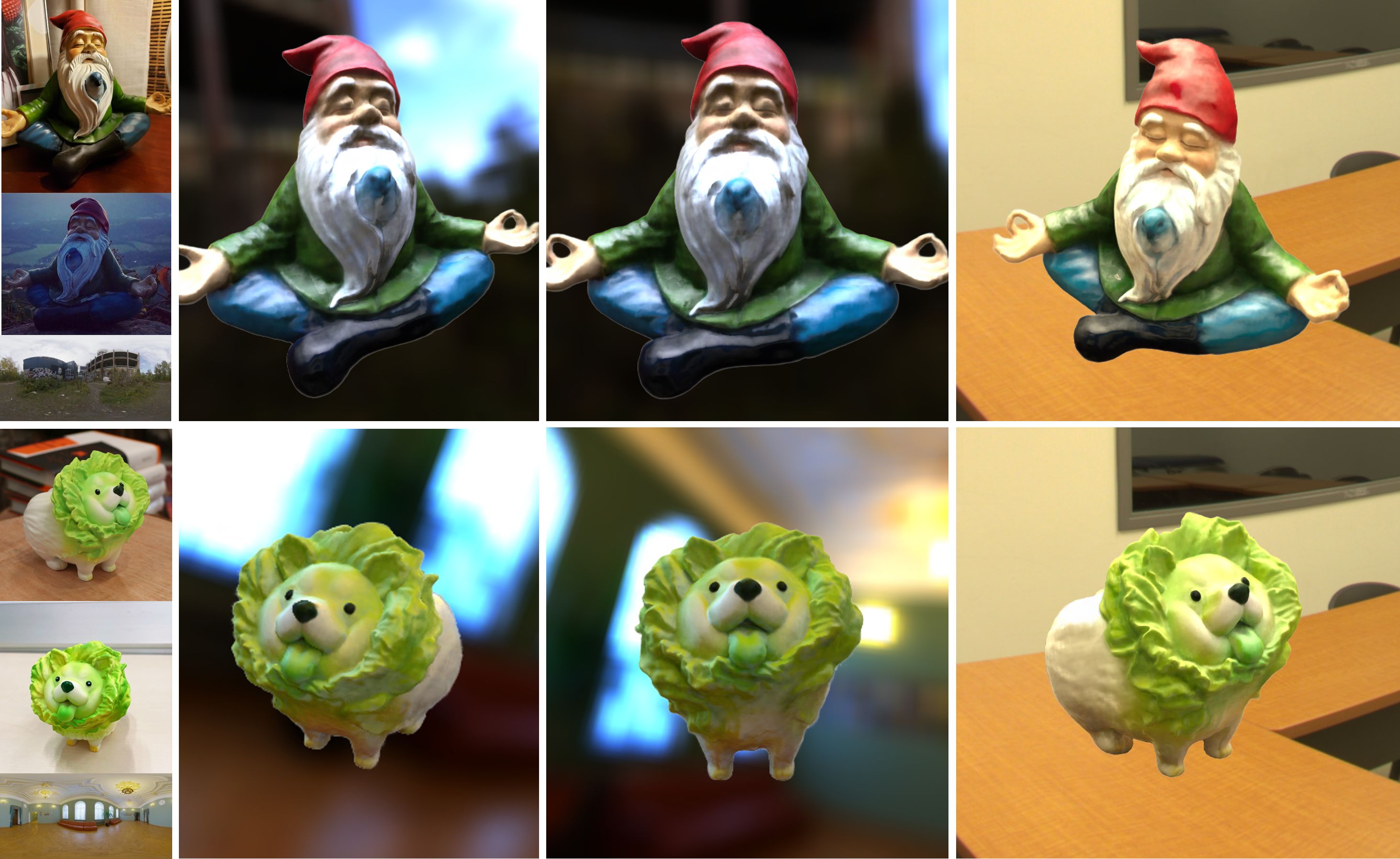}
\end{center}
\caption{
\textbf{Compositing Results.} Examples of the input online images and the environment maps are shown on the left. Our composition results are shown on the right.}
\label{fig:composite_results}
\end{figure}

Finally, we demonstrate results for our target application, rendering objects from online image collections in novel environments and lighting.
With images of several items collected from the internet, we can recover their geometry and material properties, and finally re-render and compose them into a new environment.
As shown in Fig.~\ref{fig:composite_results}, even though our input images are captured in vastly differing environments, our model handles this challenging task, producing high-quality and plausible compositing results.


\section{Limitations}
\label{sec:limitaions}

\begin{figure}[h]
\begin{center}
\newcommand{\img}[1]{\includegraphics[height=0.27\linewidth]{#1}}
\setlength{\tabcolsep}{0.5pt}
\begin{tabular}{cc|ccc}
\img{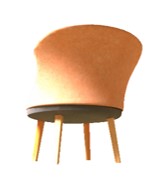} & 
\img{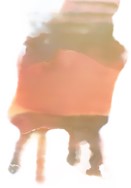} & 
\img{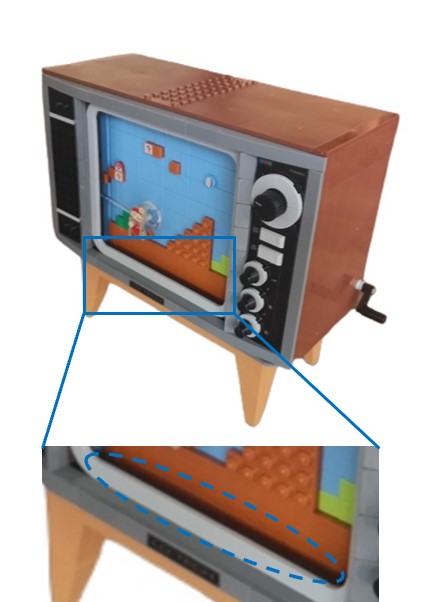} & 
\img{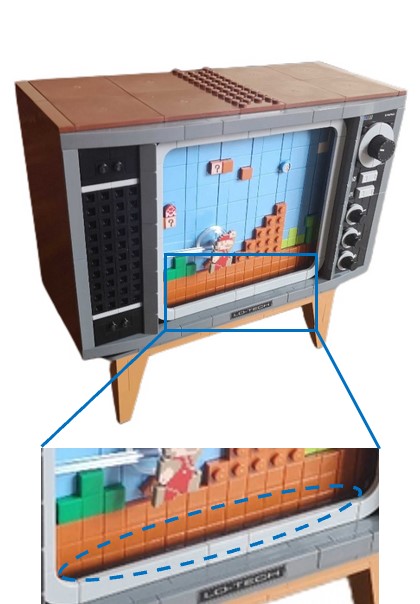} & 
\img{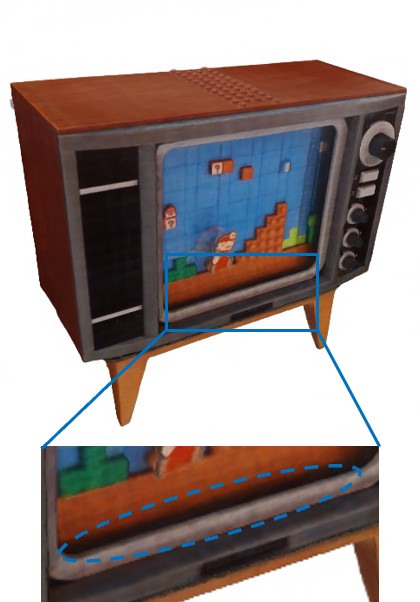} \\
{\small Ground Truth} & {\small Prediction} & {\small View \#1} & {\small View \#2} & {\small Base Color}
\end{tabular}
\end{center}
\caption{
\textbf{Failure cases of our model}: 
    \textit{Left Part}: An example of where our model failed to converge with high camera perturbations. The ground truth image is on the left and our prediction is on the right; 
    \textit{Right Part}: An example of where the ambient occlusion shading appears on the predicted base color map. From left to right are: two reference images from different views and the predicted base color map. Note that the shadow inside the grease appears both in the reference images and the base color map.}
\label{fig:limitation}
\end{figure}

As shown in Fig.~\ref{fig:limitation}, our approach has a few limitations. Firstly, while our model can handle certain camera perturbations, it still requires an initial camera pose for each training image. As shown in Tab.\ref{tab:ablation_camera}, there is a chance that our model will fail when the angle between the ground-truth poses and input poses is above 20 degrees, thus it is unlikely that our model will converge to accurate geometry with a trivial initialization of the camera poses. However, we believe this problem can be addressed by developing a more advanced camera optimization method as future work.

Although our rendering network can process complex and challenging shading in input images, \eg sharp shadows, it does not support generating such components in novel scenes. While some works~\cite{DBLP:conf/cvpr/SrinivasanDZTMB21,DBLP:journals/tog/ZhangSDDFB21} use terms such as light visibility to represent these effects in a physically-based way, applying these techniques to data with varying unknown illumination is much more challenging, and requires further investigation. In addition, some shading that appears on many training images, for example ambient occlusions, may be misinterpreted as an object's material property by our model. 
We are planning to explore these topics in our future work.

\section{Conclusion}
\label{sec:conclusion}

In this paper, we demonstrate that compelling capture, compositing, and relighting results are possible using only online image collections for which calibrated multi-view datasets captured in controlled settings are unavailable.
This opens the door for many interesting future applications, such as re-rendering objects that may one day be rare or non-extant, represented in online image collections but never explicitly captured using multi-view techniques.


\bibliographystyle{ACM-Reference-Format}
\bibliography{citation}

\clearpage
\setcounter{section}{0}
\renewcommand\thesection{\Alph{section}}

\section{Comparisons with NeRD}

In addition to the qualitative comparisons with NeRD, we also provide quantitative results in this section. We use the same training/testing dataset splits as in NeRD, and apply the same testing strategy to optimize the lighting parameters with fixed network parameters so as to fit the unknown lighting in the testing images.
The results are shown in Tab.~\ref{tab:supp_compare_NeRD}. Our model achieves better results with PSNR and competitive results with SSIM.
We also train and test NeRF~\cite{mildenhall2020nerf} on the same dataset, with the same setting as a reference. 

We also would like to further discuss the intrinsic differences between our work and NeRD:
First of all, our work aims to solve reconstruction and rendering on objects from unrestricted and multi-source online image collections.
This leads to many challenges such as inaccurate camera poses, ill-conditioned illuminations and transient occlusions, and our model is designed mainly based on how to tackle these problems.
On the other hand, NeRD only claimed to handle "image collections", and all of their experiments are conducted on data captured from single source, using the same set of cameras under similar environmental conditions.
We strongly believe our model can ourperform NeRD on datasets collected from the Internet.

From a technical standpoint, our model contains many novel features that are not present in NeRD, \ie the transient model, the expected-depth-based sampling, and the Normal Extraction Layer. Among all of these differences, we want to highlight one major difference that our renderer is based on Spherical Harmonics while NeRD uses Spherical Gaussians.
We employ Spherical Harmonics following the approach of~\citet{DBLP:conf/iccv/LinM0L21}, which jointly optimizes camera poses during NeRF training.
It proposes the idea that restricting the parameters and functions in a joint system to low-frequency components can help prevent the joint optimization from falling into local minima. As the structure of Spherical Harmonics makes it easy to control their frequency, we believe that using lower-order Spherical Harmonics can make our system more robust with unrestricted inputs.

\begin{table}[h]
\caption{\textbf{Comparison with NeRD on real datasets.} We present results in both static illumination (Cape) and varying illumination (Head, MotherChild, Gnome). We also trained the official implementation of NeRD on all our datasets. However, it did not converge on several datasets (including some of those they provided). Thus, we directly copied the results from their paper and marked them using an asterisk. We directly copy the results from their paper and mark them using an asterisk. We also report our results on NeRF here as reference. We highlight the best and second best results in each column in orange and yellow.}
\label{tab:supp_compare_NeRD}
\begin{center}
\newcolumntype{P}[1]{>{\centering\arraybackslash}p{#1}}
{\small
\begin{tabular}{|P{45pt}|P{30pt}|P{30pt}|P{30pt}|P{30pt}|}
\hline
 \multirow{2}*{Methods} & \multicolumn{2}{P{60pt}|}{Static Illum.} & \multicolumn{2}{P{60pt}|}{Varying Illum.}\\
& PSNR$\uparrow$ & SSIM$\uparrow$ & PSNR$\uparrow$ & SSIM$\uparrow$ \\
\hline
NeRF~\shortcite{mildenhall2020nerf}* & 23.34 & \btwo 0.85 & 20.11 & 0.87\\
NeRF-A* & 22.87 & 0.83 & 26.36 & \btwo 0.94\\
NeRD~\shortcite{DBLP:conf/iccv/BossBJBLL21}* & \btwo 23.86 & \bone 0.88 & 25.81 & \bone 0.95\\
NeRF & 22.95 & 0.78 & 24.31 & 0.90 \\
Ours-Geom & \bone 24.70 & 0.83 & \bone 27.96 & 0.93\\
Ours-Full & 23.47 & 0.80 & \btwo 27.16 & 0.92\\
\hline
\end{tabular}
}
\end{center}
\end{table}
\begin{figure*}[t]
\begin{center}
\includegraphics[width=\textwidth]{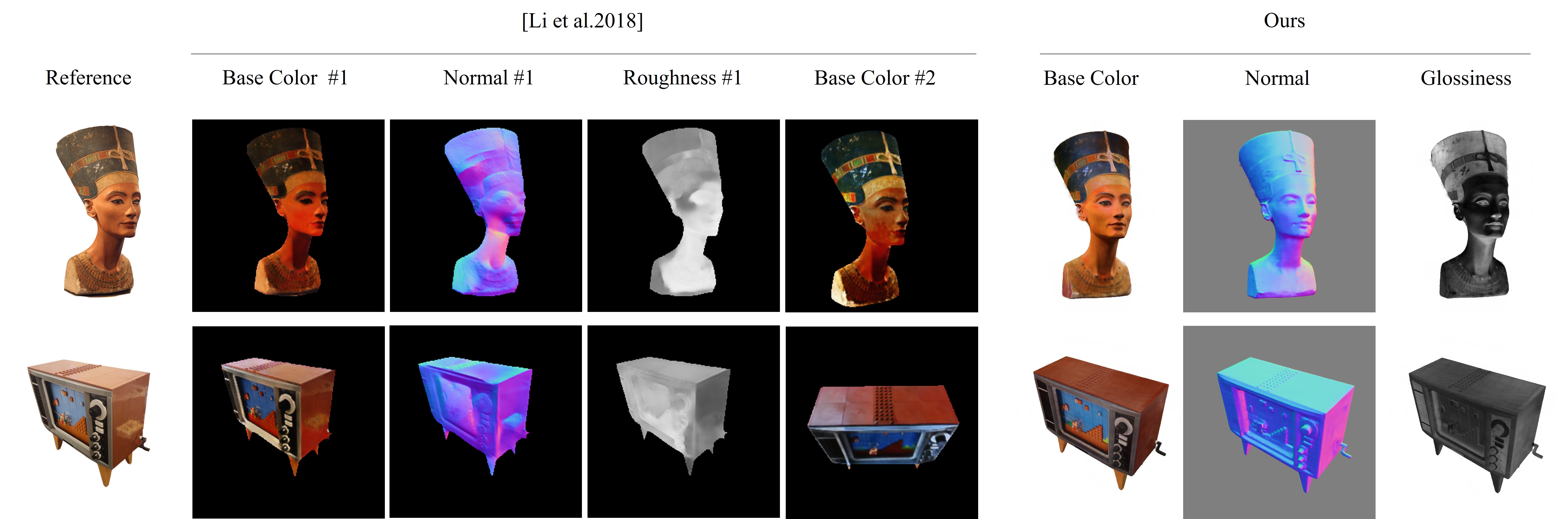}
\end{center}
\caption{
\textbf{Comparison with~\citet{li2018learning}}. On the left side, we show diffuse albedo, surface normal and roughness maps predicted by~\citet{li2018learning} from the reference image (named as \textit{Base Color \#1}, \textit{Normal \#1}, \textit{Roughness \#1}); we also show the albedo map prediction of ~\citet{li2018learning} from another input (named as \textit{Base Color \#2}). On the right side, we show the corresponding results predicted by our model. Please note that the results from our model are taken from our animated video sequences, and the full animation is in our supplementary video.}
\label{fig:qual_compare_single}
\end{figure*}

\section{Comparisons with single-image based method}
To show that our model can generate consistent and accurate material properties based on multi-view images from different lighting conditions, we compare our model with ~\citet{li2018learning}, a state-of-the-art method based on single image. As shown in Fig.~\ref{fig:qual_compare_single}, the model from~\citet{li2018learning} can generate reasonable results in some cases where the lights are relatively ambient and the object's material is diffusive, but is not capable of recovering accurate material from more challenging cases (i.e. images with mirror reflection), nor generating consistent material from different images of the same object. Our method, in the contrary, does not suffer from these challenges. It is able to learn the intrinsic and unbiased material of the object from multiple input images from different viewing angles and with variant illuminations. 

\section{Ablation Study on the sampling strategy with expected depth}
\begin{figure}[t]
\begin{center}
\newcommand{\img}[1]{\includegraphics[width=0.23\textwidth]{#1}}
\setlength{\tabcolsep}{1.0pt}
\begin{tabular}{cc}
\img{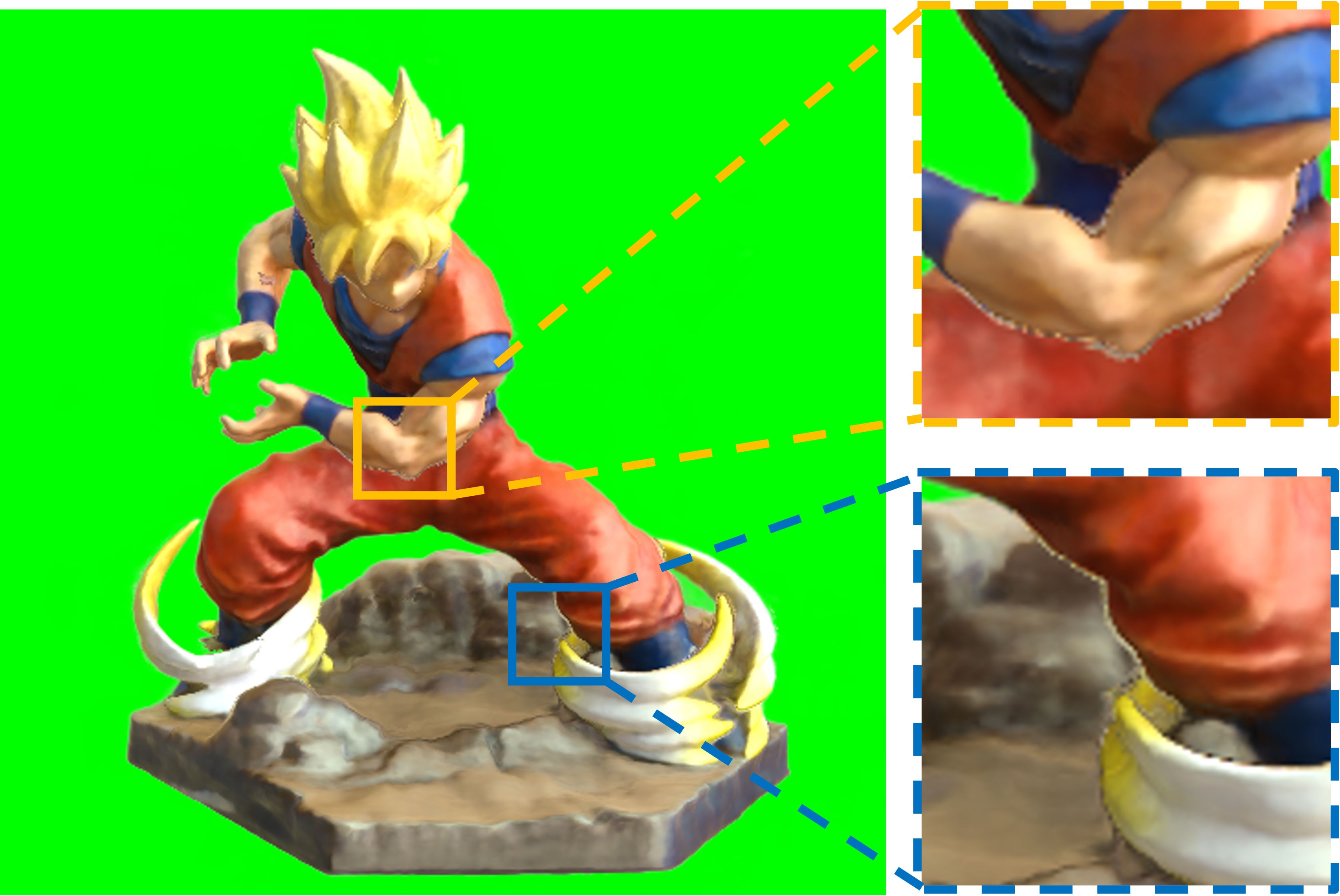} & 
\img{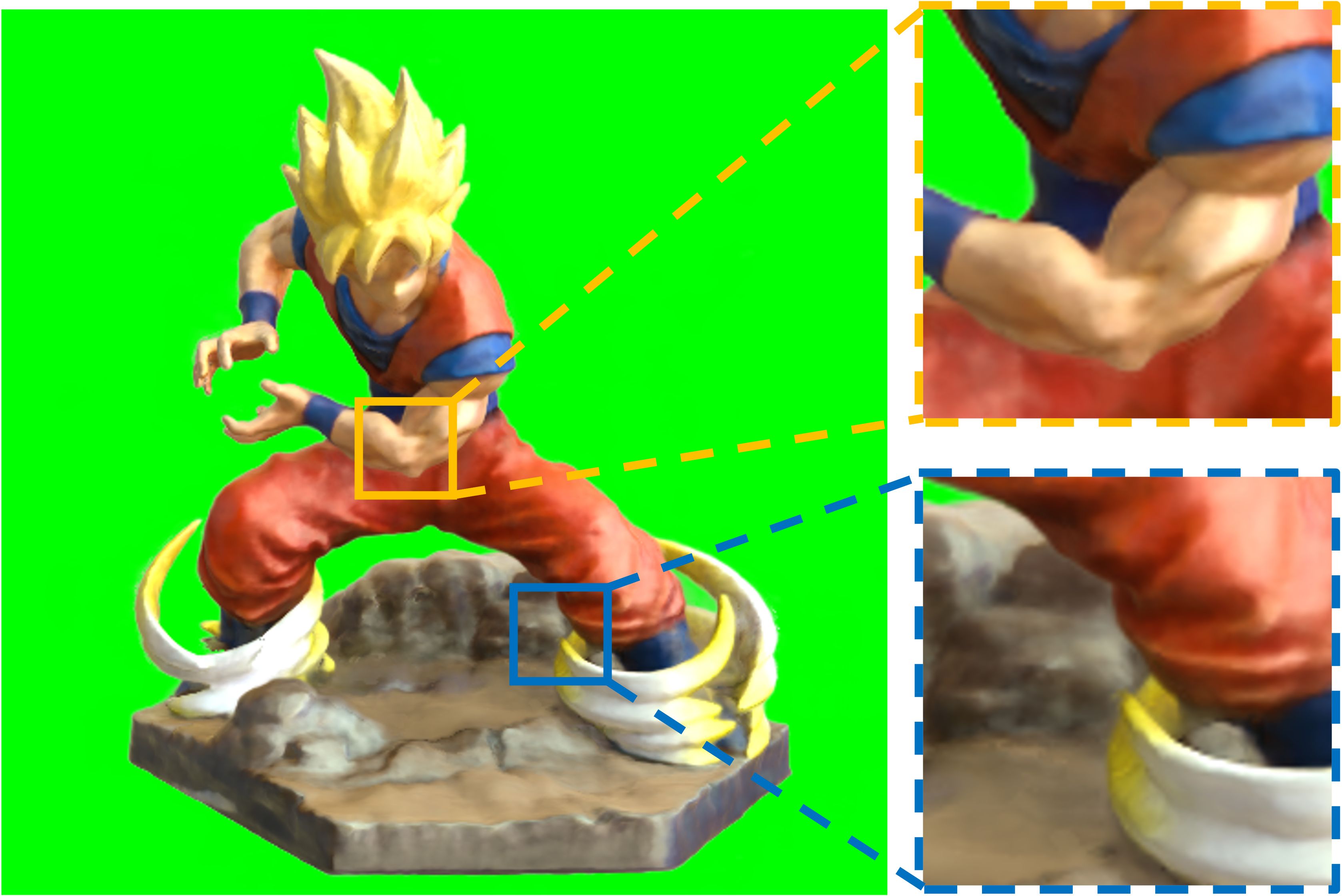} \\
\footnotesize Exp. Only (8.30 iter/s) & \footnotesize Hybrid (6.34 iter/s) \\

\img{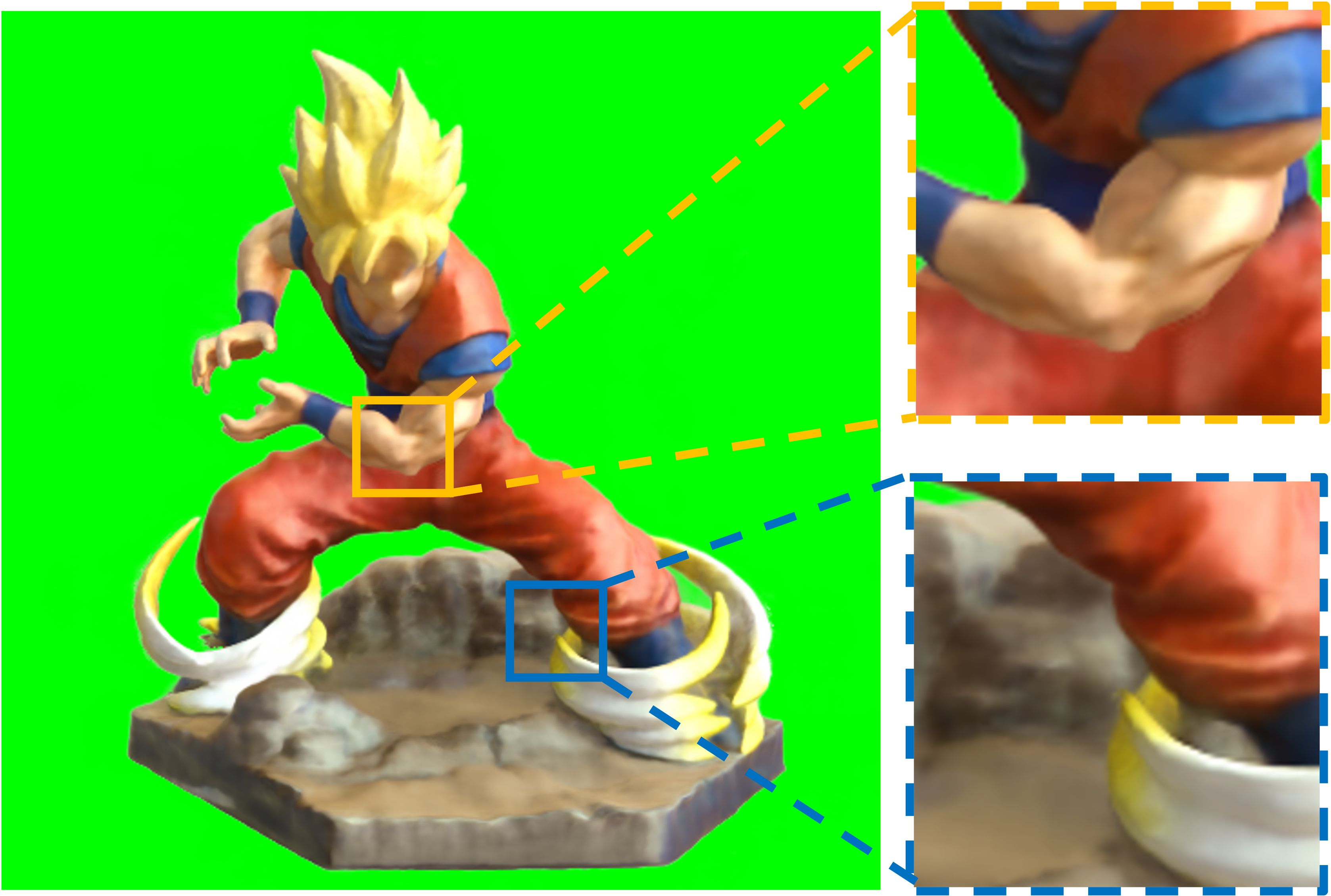} & 
\img{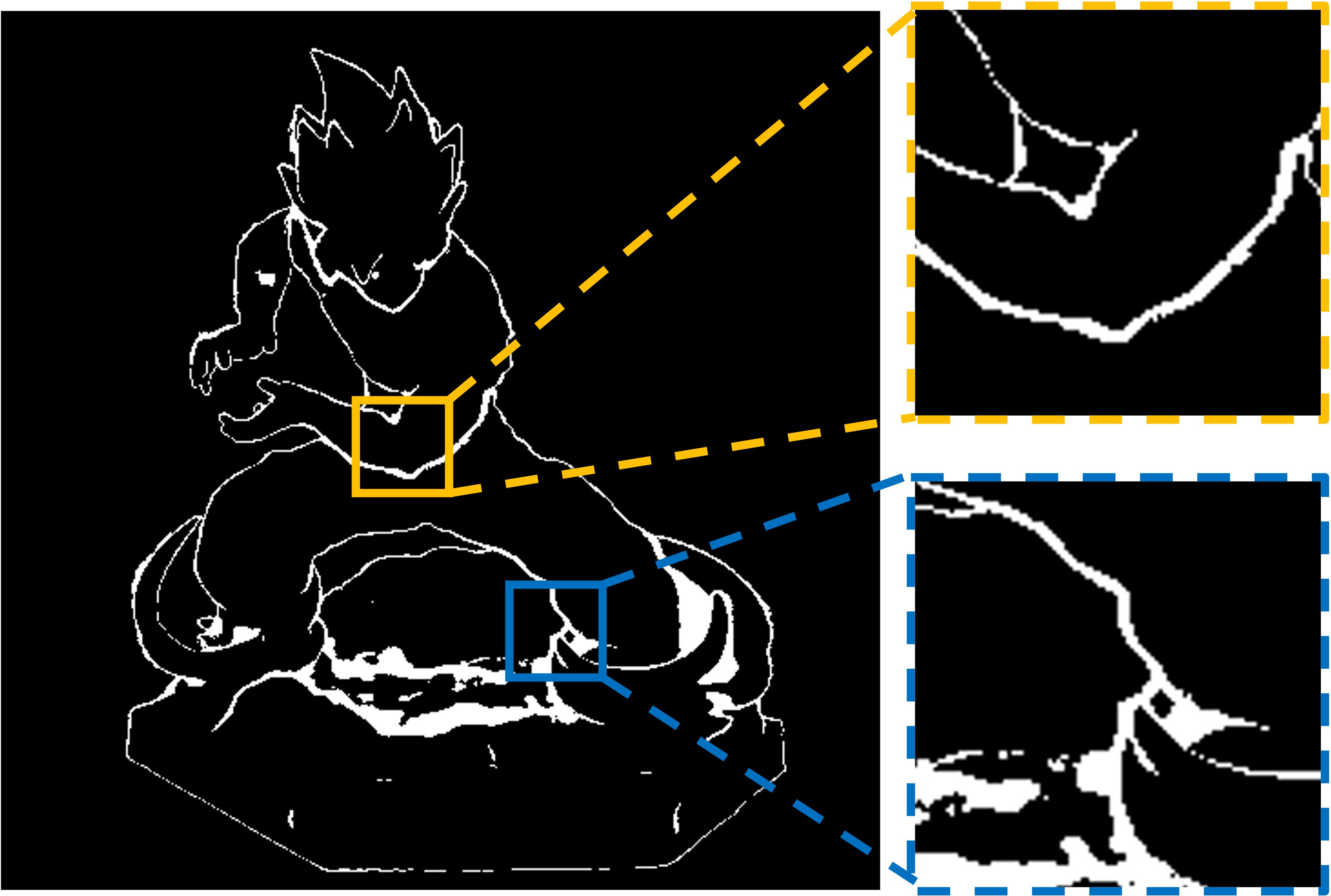} \\
\footnotesize All Points (4.62 iter/s) & \footnotesize Depth Variance Map \\

\end{tabular}
\end{center}
\caption{
\textbf{Ablation study on our sampling strategy.} From top to bottom, left to right: result with samples on expected depth only; result using hybrid sampling based on depth variance; result with all sample points; the depth variance map filtered by $\tau_d$. We also show the training speed of each model in iterations per second. }
\label{fig:ablation_exp}
\end{figure}

We show an additional ablation study on the sampling strategy with expected depth qualitatively. As shown in Fig.~\ref{fig:ablation_exp}, the hybrid method using depth variance which explained in the paper helps eliminate the aliasing effect along the object's border, while the loss of training efficiency is acceptable. We also show the depth variance map filtered by the threshold $\tau_d$, where rays from white pixels are sampled with all points and others only uses the expected depth. In all of our experiments, $\tau_d$ is set to $ (d_f-d_n) / 5000$, where $d_n, d_f$ are the depths of the near/far planes of the camera.

\begin{figure*}[t]
\begin{center}
\includegraphics[width=0.94\textwidth]{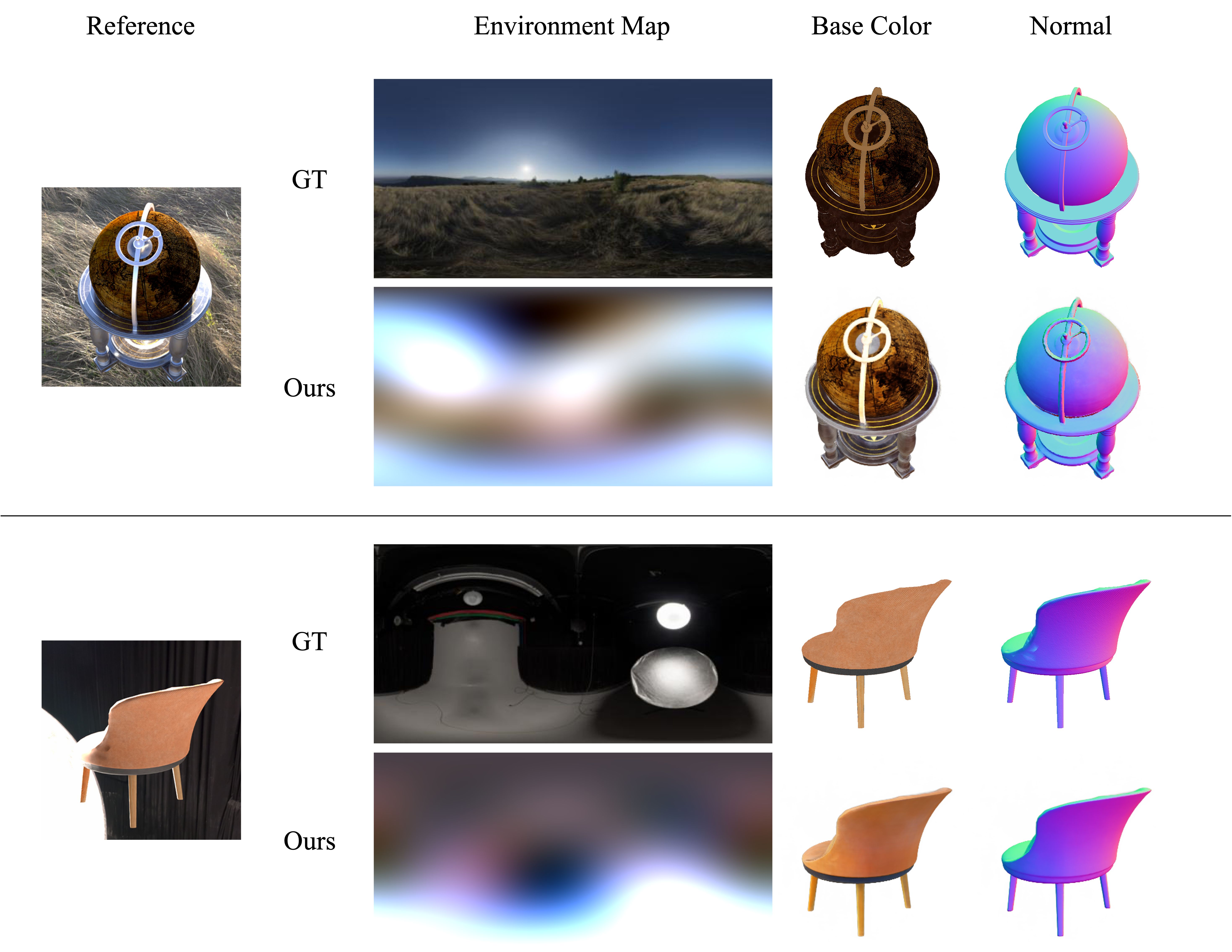}
\end{center}
\caption{
\textbf{Results on synthetic datasets}. As our BRDF model is different from the ground truth, we only show components that exist in both models, including the Base Color, Normal, and Environment Map.}
\label{fig:synthetic}
\end{figure*}

\begin{figure*}[t]
\begin{center}
\includegraphics[width=\textwidth]{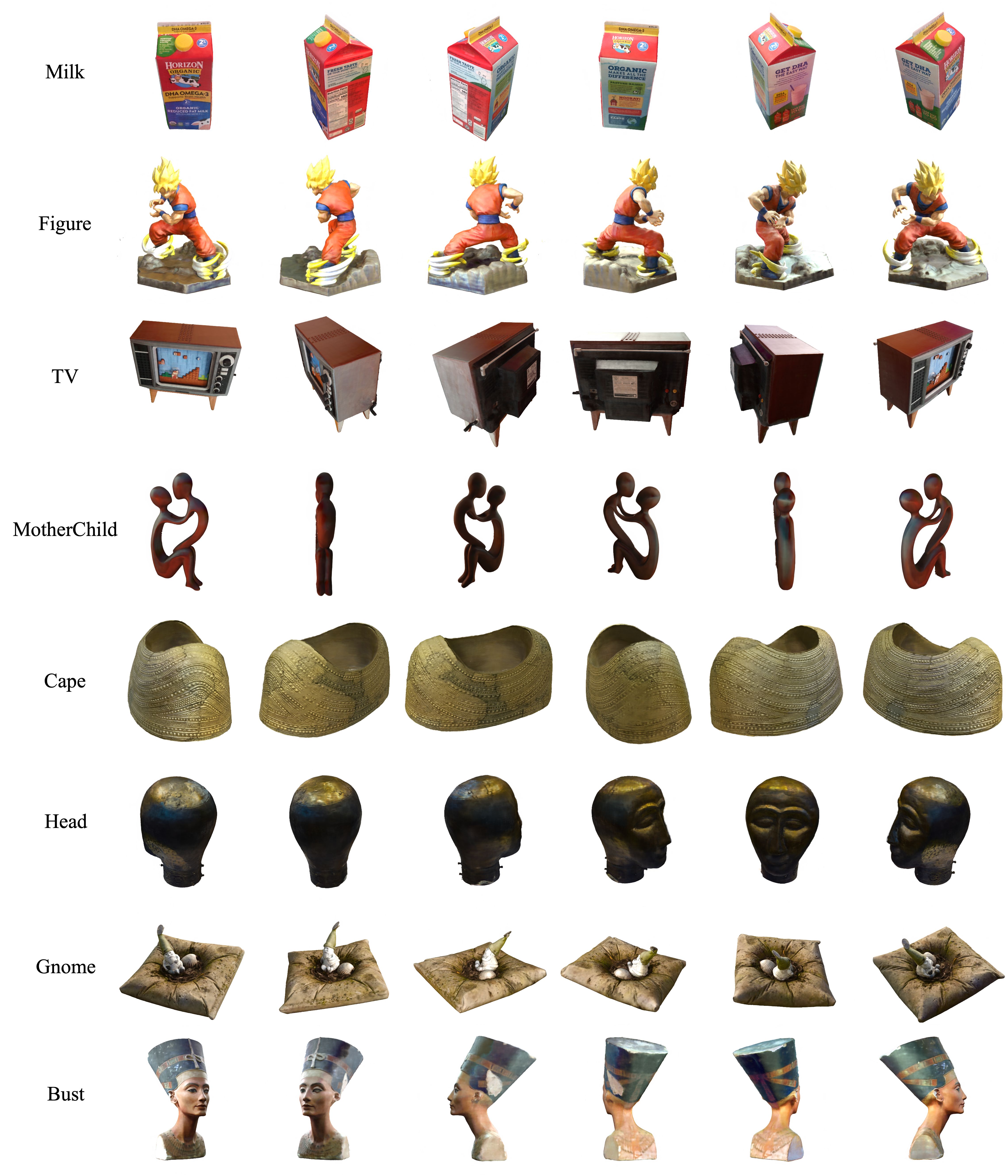}
\end{center}
\caption{
\textbf{Synthesized novel views generated using our approach}. }
\label{fig:novel_view}
\end{figure*}

\begin{figure*}[ht]
\begin{center}
\includegraphics[width=0.96\textwidth]{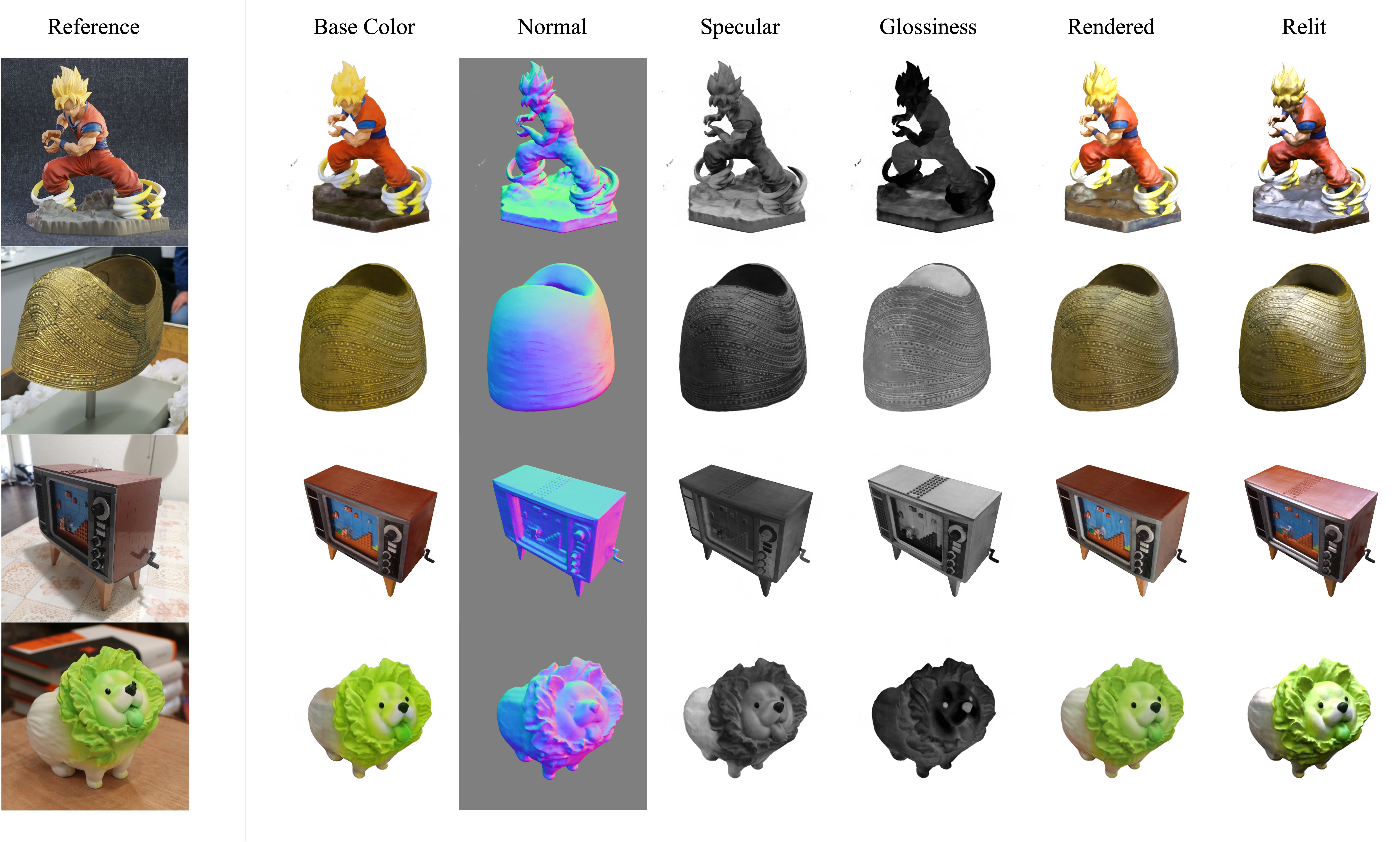}
\end{center}
\caption{
\textbf{A showcase of additional material decomposition results using our approach}. }
\label{fig:material}
\end{figure*}

\begin{figure*}[t]
\begin{center}
\includegraphics[width=0.96\textwidth]{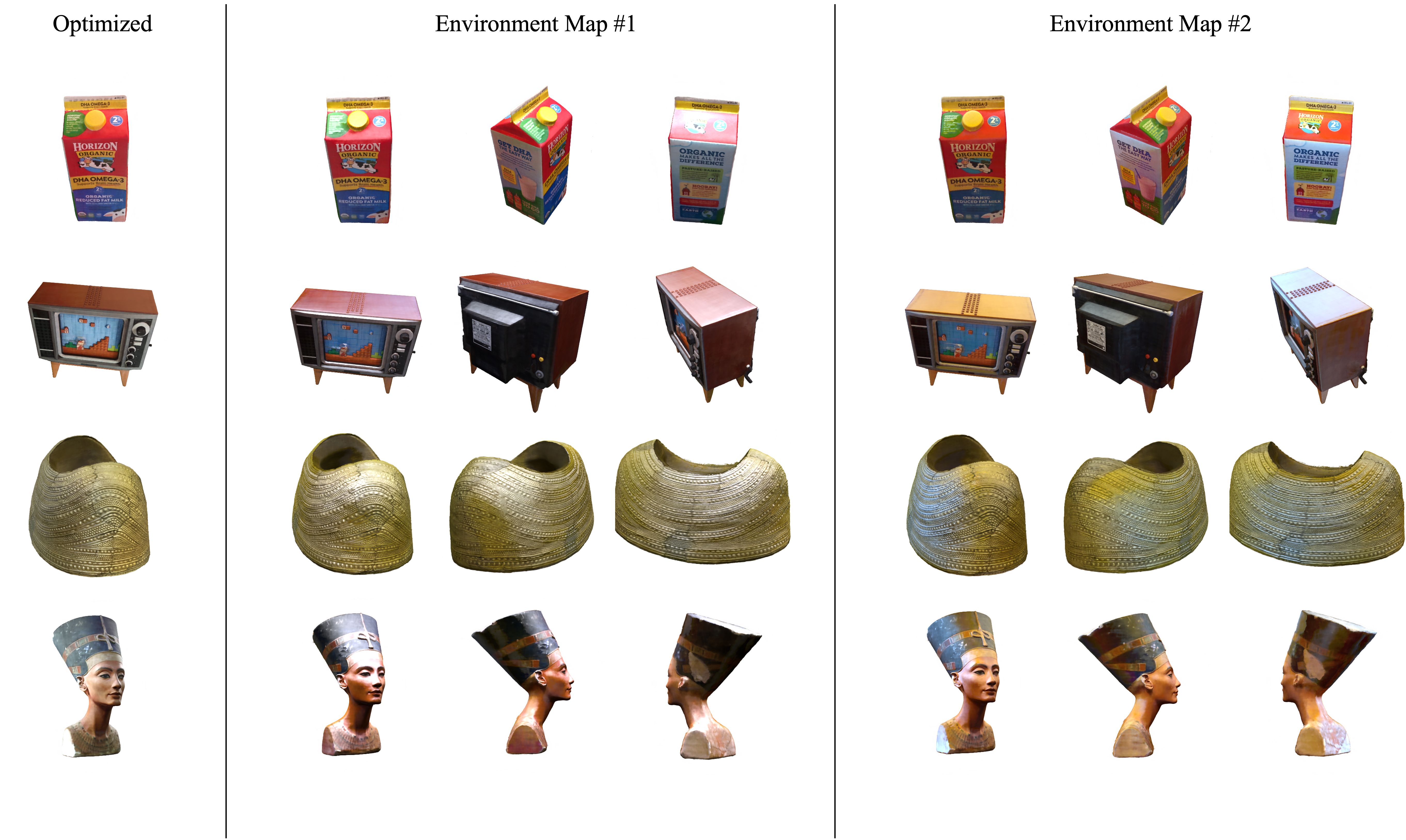}
\end{center}
\caption{
\textbf{Target objects rendered in new environments using our approach}. }
\label{fig:relight}
\end{figure*}

\section{More Results}
\label{sec:more_results}
We train our model on synthetic data to validate the quality of the material prediction. In Fig.~\ref{fig:synthetic}, we show our output materials, normal and environment maps along with the ground truth.

Even through our BRDF model and lighting parameters differ from the ground truth used when rendering these synthetic images, our method is still able to reconstruct these properties reasonably well.
This further demonstrates that our approach is robust to a wide variety of input capture or image acquisition conditions.


Lastly, we provide results from our trained model in different use cases.
Fig.~\ref{fig:novel_view} shows synthesized novel views of the target objects generated by our model; Fig.~\ref{fig:material} shows our objects' predicted materials and normals; and Fig.~\ref{fig:relight} shows rendered results of objects relit in new environments.

We note that, for the bust of Nefertiti seen in Figs.~\ref{fig:novel_view} and~\ref{fig:relight} (last columns), capturing our own input images for object capture would be infeasible, given the manner in which this object is stored in a remote and secure location.

We also provide animated results for these objects in our supplementary video.

\section{Implementation Details}
\subsection{SH Rendering Model}
Spherical Harmonics (SH) represents a group of basis functions defined on the sphere surface, commonly used for factorizing functions and fast integration for multiplying functions. A Spherical Harmonic $ Y_{lm}(\theta, \phi) $ of index $l, m$ is defined as:
\begin{equation}
    Y_{lm}(\theta, \phi) = \sqrt{\frac{2l+1}{4\pi}\frac{(l-m)!}{(l+m)!}}P_l^m(\cos \theta)e^{Im\phi},
\end{equation}
where $0\leq l\leq+\infty,-l\leq m\leq l$, and $P_l^m(\cos \theta)e^{Im\phi}$ are the associated Legendre polynomials.

Below we describe our rendering pipeline using Spherical Harmonics. Our model aims to calculate the single-bounce light reflections on the object surface from a spherical environment map $L$, where the light transport equation is defined as:
\begin{equation}
    B(\bm{n}, \bm{\omega}_o) = \int_{\bm{\omega}_i \in \Omega^+} L(\bm{\omega}_i) \rho(\bm{\omega}_o, \bm{\omega}_i) (\bm{n}\cdot \bm{\omega}_i) d\bm{\omega}_i,
\end{equation}
where $\bm{n},\bm{\omega}_i,\bm{\omega}_o$ are directions of surface normal, incoming light, and outgoing light, $\Omega^+$ is the upper hemisphere above the surface, and $B(\bm{n},\bm{\omega}_o),L(\bm{\omega}_i),\rho(\bm{\omega}_i, \bm{\omega}_o)$ are the outgoing light towards direction $\bm{\omega}_o$, the incoming light from direction $\bm{\omega}_i$, and the bidirectional reflectance distribution function (BRDF) between $\bm{\omega}_i$ and $\bm{\omega}_o$, respectively.

According to~\citet{sh-rendering}, functions $L$ and $\rho$ can be approximated by a group of SHs $Y_{lm}(\bm{\omega})$ as:
\begin{align}
    L(\bm{\omega}_i) &\approx \sum_{l,m} L_{lm} Y_{lm}(\bm{\omega}_i), \\ 
    \rho(\bm{\omega}_i, \bm{\omega}_o) &\approx \sum_{l,m}\sum_{p,q} \rho_{lm,pq} Y^*_{lm}(\bm{\omega}_i) Y_{pq}(\bm{\omega}_o),
\end{align}
where $0\leq\{l,p\}\leq +\infty,-l\leq m\leq l,-p\leq q\leq p$, $Y^*_{lm}$ is the conjugate of $Y_{lm}$, and $L_{lm}$, $\rho_{lm,pq}$ are coefficients calculated by applying an integration on the multiplication of functions $L$, $\rho$ and the SHs.

If the BRDF $\rho$ is isotropic, we can reduce its number of coefficient indices to three, denoted as $\rho_{lpq}$. The outgoing light field $B$ can thus be approximated as:
\begin{equation}
B(\bm{n},\bm{\omega}_o)\approx\sum_{l,m,p,q}B_{lmpq}C_{lmpq}(\bm{n},\bm{\omega}_o),
\label{eq:B_hard}
\end{equation}
where $B_{lmpq}=\Lambda_lL_{lm}\rho_{lpq}$, $\Lambda_l=\sqrt{4\pi/(2l+1)}$ is a normalizing constant, and $C_{lmpq}(\bm{n},\bm{\omega}_o)$ is a set of basis functions which are not discussed in detail here.

If the BRDF is independent of $\bm{\omega}_o$, we can further simplify Eq.~\ref{eq:B_hard} by removing $\bm{\omega}_o$ as:
\begin{equation}
B(\bm{n})\approx\sum_{l,m}B_{lm}Y_{lm}(\bm{n}),
\label{eq:B_easy}
\end{equation}
where $B_{lm}=\Lambda_l L_{lm}\rho_{l00} \: \dot= \: \Lambda_l L_{lm}\rho_{l}$. 

We use the Phong BRDF model~\cite{phong1975illumination} to represent the object material in all our experiments, which is defined as:
\begin{equation}
    \rho(\bm{\omega}_i,\bm{\omega}_o)=\frac{K_d}{\pi}(\bm{\omega}_i\cdot\bm{n})+\frac{K_s(g+1)}{2\pi}(\bm{\omega}_i\cdot\bm{\omega}_r)^g,
    \label{eq:phong}
\end{equation}
where $K_d,K_s,g$ are parameters of the base color, specularity, and glossiness, and $\bm{\omega}_r$ is the reflection of $\bm{\omega}_o$. We calculate the two terms in Eq.~\ref{eq:phong} separately.

The first term is also known as the Lambertian BRDF.
It has been demonstrated that calculating Eq.~\ref{eq:B_easy} with $l<=2$ can capture more than $99\%$ of the reflected radiance of this term.
Let $A_l=\Lambda_l \rho_l$ be the normalized coefficient of term $(\bm{\omega}_i \cdot\bm{n})$, we have $A_0=3.14,A_1=2.09,A_2=0.79$. 
Bringing them into Eq.~\ref{eq:B_easy}, we can calculate the Lambertian term by querying the value of each SH at $\bm{n}$, calculating the weighted sum, and finally multiplying it with $K_d/\pi$.

As for the second term, we adopt the method from~\citet{sh-rendering}, in which we replace $\bm{n}$ with $\bm{\omega}_r$ in Eq.~\ref{eq:B_hard}, thus making it independent of $\bm{\omega}_o$ and reducible to Eq.~\ref{eq:B_easy}.
In this case, the approximation of the BRDF coefficients is given as:
\begin{equation}
    \Lambda_l \rho_l \approx \exp(-\frac{l^2}{2g}).
\end{equation}
The remaining steps are then the same as the first term.

Our renderer is implemented in PyTorch~\cite{NEURIPS2019_9015} and is fully differentiable.
In all our experiments, we set $l\leq 3$, which leads to 16 light coefficients $L_{lm}$ for each color channel to optimize (in total $16\times 3 = 48$ parameters).
Parameters $K_d,K_s$ are limited to $[0,1]$, and $g\in [1, +\infty]$.
To reduce the ambiguity, we assume white specular highlights, and thus setting the channels of $K_s$ to 1.

\subsection{Losses}
Here we explain the losses in more details.
The color reconstruction loss $\mathcal{L}_\text{c}$ and the transient regularity loss $\mathcal{L}_\text{tr}$ introduced in Sec.~3.3 in the paper are defined as:
\begin{align}
    \mathcal{L}_\text{c}(\bm{r}) & = \frac{\lVert C_k(\bm{r}) - \mathcal{I}_k(\bm{r})\rVert_2^2}{2\beta_k(\bm{r})^2} + \frac{\log (\beta_k(\bm{r})^2)}{2}, \\
    \mathcal{L}_\text{tr}(\bm{r}) & = \frac{1}{N_p}\sum_{i=1}^{N_p}\sigma^{(\tau)}_k(\bm{x}_i),
\end{align}
where $\bm{r}$ is a ray from image $\mathcal{I}_k$ and $\bm{x}_i$ are the sample points along $\bm{r}$. $\beta_k(\bm{r})$ is the uncertainty along the ray $\bm{r}$, which integrates the uncertainty predictions at all sample points.

\subsection{Dataset \& Training Details}

\begin{table}[t]
  \caption{\textbf{Details of our datasets.} We split our datasets into three categories based on their sources (from~\citet{DBLP:conf/iccv/BossBJBLL21}, self-captured, and collected from the Internet).}
\label{tab:dataset}
\begin{center}
\newcolumntype{P}[1]{>{\centering\arraybackslash}p{#1}}
{\small
\begin{tabular}{|c|P{40pt}|P{40pt}|P{40pt}|c|}
\hline 

Dataset & Image \# & Train \# & Test \# & $\lambda$ in NEL \\

\hline
\multicolumn{5}{|c|}{From~\citet{DBLP:conf/iccv/BossBJBLL21}} \\
\hline 

Cape & 119 & 111 & 8 & 1 \\
Head & 66 & 62 & 4 & 1 \\
Gnome & 103 & 96 & 7 & 0.1 \\
MotherChild & 104 & 97 & 7 & 1 \\
Chair & 260 & 250 & 10 & 1 \\
Car & 260 & 250 & 10 & 1 \\
Globe & 260 & 250 & 10 & 1 \\

\hline 
\multicolumn{5}{|c|}{Self-Captured} \\
\hline 

Figure & 49 & 43 & 6 & 0.1 \\
Milk & 43 & 37 & 6 & 1 \\
TV & 40 & 35 & 5 & 1 \\

\hline
\multicolumn{5}{|c|}{From the Internet} \\
\hline 

Gnome2 & 35 & 32 & 3 & 1 \\
Dog & 36 & 33 & 3 & 1 \\
Bust & 41 & 38 & 3 & 1 \\

\hline 
\end{tabular}
}
\end{center}
\end{table}

As mentioned in our paper, our datasets are from three sources:  Image Collection from the internet; Objects we captured ourselves; and Realistic data published in NeRD~\cite{DBLP:conf/iccv/BossBJBLL21}. For the first part, we downloaded user review images of two popular products (named as \textit{Dog} and \textit{Gnome2} in the following context) on Amazon and Taobao, two of the most widely-used online shopping websites, and online images of the Nefertiti (named as \textit{Bust}) from the Google image searching engine; For our self-captured data, we captured 3 objects (\textit{Milk}, \textit{Figure}, \textit{TV}) using our own devices, where each object is placed in about 4-6 different scenes. Finally, we obtained the dataset from NeRD's project website and used 4 of its real object image collections (\textit{Gnome}, \textit{Head}, \textit{Cape}, \textit{MotherChild}) and 3 of its synthetic collections (\textit{Chair}, \textit{Car}, \textit{Globe}) in our evaluation. 

For all of the self-collecting datasets, approximately 40 images are collected for each object. Then, we use the SfM pipeline in COLMAP~\cite{colmap} to register the initial camera poses, with image matches generated from SuperGlue~\cite{superglue}. The foreground masks are calculated using the online mask extraction pipeline of remove.bg~\cite{removebg}.

Tab.~\ref{tab:dataset} lists the numbers of images and configurations of our datasets.
Since the controllable parameter $\lambda$ in our normal extraction layer (NEL) is not fixed for all scenes, we also list its values in the rightmost column of the table. 

We generate and store rays for all pixels from the input image before training starts.
At the beginning of each epoch, we use the foreground masks to ensure that the number of the chosen background rays does not exceed the foreground rays by more than a factor of $2$, and then concatenate and shuffle the background and foreground rays together.

In the first stage, we decay the learning rate by a factor of $0.3$ at intervals of $10$ epochs. In the second stage, we use the cosine annealing schedule~\cite{DBLP:conf/iclr/LoshchilovH17} with $T_\text{max}=10$ to reduce the learning rate, as the training epoch is relatively small.

Since the SfM pipeline of COLMAP also produces a sparse point cloud of the target object while solving camera poses, we further use them to help train our model. We generate a coarse bounding box of the object based on the points, and only sample ray points inside the bounding box. In contrast, the data from~\citet{DBLP:conf/iccv/BossBJBLL21} are captured in the same scene, and the background is also used in their camera registration, making this optimization infeasible in their approach.


\end{document}